\documentclass{article}

\usepackage{graphicx}
\usepackage{apacite}
\usepackage{arxiv}
\usepackage[utf8]{inputenc} % allow utf-8 input
\usepackage[T1]{fontenc}    % use 8-bit T1 fonts
\usepackage{url}            % simple URL typesetting
\usepackage{booktabs}       % professional-quality tables
\usepackage{amsfonts,amsmath}       % blackboard math symbols
\usepackage{nicefrac}       % compact symbols for 1/2, etc.
\usepackage{microtype}      % microtypography
\usepackage{algorithm}
\usepackage[noend]{algpseudocode}
\usepackage{subcaption}
\usepackage{hyperref}
\usepackage{epstopdf}
\usepackage{amsthm}
\usepackage{multicol}
\usepackage{natbib}
\usepackage{wrapfig}
\usepackage{bbm}

\hypersetup{
    colorlinks=true,
    linkcolor=red,
    filecolor=magenta,      
    urlcolor=cyan,
}

\DeclareMathOperator{\EX}{\mathbb{E}}
\newcommand{\argmax}{\mathop{\mathrm{argmax}}}

\newcommand{\BlackBox}{\rule{1.5ex}{1.5ex}}  % end of proof
\ifdefined\proof
    \renewenvironment{proof}{\par\noindent{\bf Proof\ }}{\hfill\BlackBox\\[2mm]}
\else
    \newenvironment{proof}{\par\noindent{\bf Proof\ }}{\hfill\BlackBox\\[2mm]}
\fi

\newtheorem{theorem}{Theorem}
\newtheorem{lemma}{Lemma} 
 
\newtheorem{remark}{Remark}

\author{
  Alexander Galozy\\
  Center for Applied Intelligent Systems Research\\
  Halmstad University\\
  Kristian IV:s väg 3, 301 18 Halmstad \\
  \texttt{alexander.galozy@hh.se} \\
   \And
 S\l{}awomir Nowaczyk \\
  Center for Applied Intelligent Systems Research\\
  Halmstad University\\
  Kristian IV:s väg 3, 301 18 Halmstad \\
  \texttt{slawomir.nowaczyk@hh.se} \\
  \AND
   Mattias Ohlsson \\
Center for Applied Intelligent Systems Research\\
  Halmstad University\\
  Kristian IV:s väg 3, 301 18 Halmstad \\
  \texttt{mattias.ohlsson@hh.se} \\
}



\begin{document}
\title{A New Bandit Setting Balancing Information from\\ State Evolution and Corrupted Context}
% several? different? diverse? competing?

% A New Bandit Setting Balancing Information on State Evolution and Corrupted Context?
% Action Sequences? 

\maketitle
\begin{abstract}%
We propose a new sequential decision-making setting, combining key aspects of two established online learning problems with bandit feedback. The optimal action to play at any given moment is contingent on an underlying changing state which is not directly observable by the agent. Each state is associated with a context distribution, possibly corrupted, allowing the agent to identify the state. Furthermore, states evolve in a Markovian fashion, providing useful information to estimate the current state via state history. In the proposed problem setting, we tackle the challenge of deciding on which of the two sources of information the agent should base its arm selection. We present an algorithm that uses a referee to dynamically combine the policies of a contextual bandit and a multi-armed bandit. We capture the time-correlation of states through iteratively learning the action-reward transition model, allowing for efficient exploration of actions. Our setting is motivated by adaptive mobile health (mHealth) interventions. Users transition through different, time-correlated, but only partially
observable internal states, determining their current needs. The side information associated with each internal state might not always be reliable, and standard approaches solely rely on the context risk of incurring high regret. Similarly, some users might exhibit weaker correlations between subsequent states,
leading to approaches that solely rely on state transitions risking the same. We analyze our setting and algorithm in terms of regret lower bound and upper bounds and evaluate our method
on simulated medication adherence intervention data and several real-world data sets, showing
improved empirical performance compared to several popular algorithms.
\end{abstract}
% keywords can be removed


\section{Introduction}
In the area of sequential decision-making, the multi-armed bandit problem has attracted significant attention due to its applicability in many real-world areas such as clinical trails~\citep{Villar2015, Bastani2020}, finance~\citep{Shen2015, Huo2017}, routing networks~\citep{Boldrini2018, Kerkouche2018}, online advertising~\citep{Wen2017, Schwartz2017} and movie~\citep{Wang2019} or app recommendation~\citep{Baltrunas2015}. The agent's goal is to select an action from a set of available actions (also known as arms) that would minimize the regret, commonly defined as the difference between the cumulative rewards of the agent's strategy and an a priori known best strategy (i.e., one that chooses the action with the highest expected reward at each decision point). 

The earliest works have formalized this problem in the so-called Multi-Armed Bandit ($MAB$) problem. In the stochastic variant of the $MAB$ problem~\citep{Lai1985}, the rewards the agent receives by playing an action are a sequence of i.i.d. random variables. Popular methods for the design of action selection strategies include using upper-confidence-bounds (UCB) on the mean rewards of actions, based on the principle of "optimism in the face of uncertainty"~\citep{Lai1985, Auer2002}. Other approaches rely on the Bayesian interpretation for optimal exploration and exploitation using Thompson Sampling~\citep{Thompson1933, Chapelle2011}. For Bernoulli distributed rewards, Thompson Sampling has been shown to outperform state-of-the-art methods that use UCB-type strategies and achieve finite-time regret guarantees that match the asymptotic rate of the lower bound for this setting~\citep{Kaufmann2012}. Indeed, even in the case of Gaussian distributed rewards, Thompson Sampling enjoys a near-optimal upper bound that matches the lower bound~\citep{Agrawal2013}. 

This paper is motivated by mobile health (mHealth), mainly by providing tailored digital interventions to promote behavior change. To model this setting, we assume that an underlying evolving state determines users' needs and wants. Thus, we expect non-stationarity in the rewards of interventions so that different interventions will be beneficial depending on the state. For example, some patients are affected by stress and require stress-coping techniques. In contrast, others may experience little stress but require information and knowledge instead~\citep{Reda1994}. The underlying state, as it evolves, may induce a natural ordering of interventions as the patient transitions through different ''stages''. For example, according to the transtheoretical model, patients often go through stages to enact a change in their habits. The ''Preparation'' stage is characterized by the willingness to change behaviors, where small steps are taken towards the goal. In the ''Maintenance'' stage, patients have changed their habits and are committed to preventing relapse into earlier stages~\citep{prochaska2015}. While assuming intervention will affect the state of the user, we do not expect this effect to be drastic. Behavior change is a long process, where any single intervention is unlikely to have a significant influence on the users~\citep{Huitian2017}. Thus, modeling the task of providing tailored interventions via the bandit framework becomes a valid alternative to full reinforcement learning, providing good sample complexity and provable regret guarantees. 

The stochastic $MAB$ problem assumes that the average rewards of actions do not change over time, that is, reward distributions are \textit{stationary}. While a convenient and straightforward model of online decision-making, this formulation is quite limiting in many interesting practical settings, where the mean reward of actions may change over time. For example, a user's taste in a particular movie genre might shift, or treatment might reduce effectiveness with repeated exposure. Consequently, modeling our problem using a $MAB$ formulation would not fit particularly well with our assumptions, and a non-stationary approach is required. Various non-stationary settings have been explored, such as switching bandits. In switching bandits, mean rewards of actions may change abruptly or continuously over time. For stochastic rewards and a fixed number of distribution changes, authors in~\cite{Garivier2011} prove regret lower bound of $\Omega(\sqrt{T})$. Their analysis assumes that the reward distribution changes independently for each action, such that a player may miss the instance where another arm may become the best action. For the agent to achieve sublinear regret, the number of distribution changes needs to be of the order less than $T$. Otherwise, the agent may continue to choose sub-optimal actions. Authors in~\cite{Auer2003b} show that their EXP3.S algorithm achieves $O\big(\sqrt{KLT}\big)$ regret, where $L$ is the number of distribution changes known in advance. Without knowing the number of distribution changes in advance,~\cite{Yu2009} provide an algorithm that achieves dynamic regret of order $O\big(KL \log(T)\big)$. Recently it has been shown that for the stochastic non-stationary bandit problem, achieving optimal regret is possible for the two-armed bandit problem without knowing $L$ in advance, where the authors present an algorithm that achieves the optimal lower bound of order $O\big(\sqrt{KLT}\big)$~\citep{Auer19a}. These prior works focus on independent changes between arms and do not quite reflect our assumptions. In our setting, arms are correlated such that the reward distributions of arms change \textit{simultaneously}. Therefore, detecting a change point may become significantly easier.

Our problem setting exhibits more structure in how the changes in reward distributions evolve. The most closely related work focuses on so-called \textit{rested}~\citep{Gittins1979} and \textit{restless} bandits~\citep{Whittle88}, particularly when the state dynamics of actions are Markovian defined in the restless Markov bandits problem explored by~\cite{Ortner2014}. For their problem setting, they prove a regret lower bound of $\Omega\big(\sqrt {ST}\big)$, where $S:=Km$ with $K$ being the number of actions and $m \geq 1$. What separates these problems from ours is a simple restriction: We do not have an independent Markov process for each action determining the action's reward distribution, but a common Markov process. A similar setting to ours has been explored by authors in \textit{regime-switching bandits}~\citep{Zhou2021}. They compare performance to an optimal partially observable Markov decision process (POMDP) that knows the transition matrix and reward distribution, and their results in terms of regret are not directly applicable to our setting. The regime-switching bandit setting assumes that the hidden state is not accessible to the agent outside of action rewards. However, additional data can often be collected to estimate it. For example, questionnaires or passively collected data such as geolocation and physical activity metrics. Thus, while a better fit for mHealth, regime-switching bandits do not capture this crucial property of our problem setting. 

One particular bandit setting that exploits additional information has received significant attention in contemporary works. This extension to the $MAB$-problem is the Contextual Bandit ($CB$) problem, also known as the \textit{multi-armed-bandit problem with side information} or \textit{associative reinforcement learning}. Additional information at each time step effectively turns the non-stationary problem into a stationary one. This additional information is often referred to as \textit{context} in the bandit literature, which the agent uses to select actions. The goal of the agent, then, is to learn the relationship between context and the action rewards, allowing the prediction of action rewards and thus choosing the optimal one. The $CB$-agent has a distinct advantage over $MAB$-agents in the $CB$ setting: while the $CB$-agent can use the context to ``detect'' changes in reward distribution (or state), the $MAB$-agent has to discover changes through the rewards it receives. For the contextual bandit setting with a linear context to reward mapping, ~\cite{Chu2011} prove a regret lower bound of order $\sqrt{Td}$, with $d$ being the dimensionality of the context and $T$ being the number of time steps. Various algorithms have been developed that achieve regret of the lower bound. Some notable works for the contextual bandit problem include Linear Upper-Confidence-Bound (LinUCB)~\citep{Chu2011, Li2010} and Linear Contextual Thompson Sampling (LinTS)~\citep{agrawal2014thompson}.

Naturally, the acquired context does not, in practice, give a complete picture and may be susceptible to noise corruption. Standard contextual bandit approaches discussed before may fail to achieve the regret guarantees due to noisy contexts.~\cite{bouneffouf2017context} investigated the problem setting they call \textit{contextual bandits with the restricted context}. In this setting, the authors deal with feature sparsity, where only a small subset of features is relevant for reward prediction. The superfluous variables introduce noise and delay learning of the context to action mapping, incurring additional regret. Their approach does not address the issue of complete corruption, where whole contexts can be rendered useless for decision making. Another work related to context noise by~\cite{Yun2017} investigate the CB-problem under feature uncertainty. They assume features are corrupted by random i.i.d. noise on features, but do not address the problem of arbitrary corruption. While an important first step, the mechanism of context corruption is not expected to be governed by random i.i.d distributed noise in mHealth. It is, therefore, questionable whether the true reward model is sufficiently learnable when relying on such an assumption. We expect users not to know their state perfectly or pretend for various reasons and thus provide us with context that is a noisy estimate of their underlying state. To equip our problem setting with the property of uncertain contexts, we consider a bandit problem introduced by~\cite{Bouneffouf2020}, referred to as the contextual bandit with corrupted context, where the agent may not be able to use the provided contexts due to significant corruption. 

We incorporate non-stationarity in the rewards and context corruption in our problem setting, like in regime-switching bandits, a hidden state that evolves according to a Markov chain and determines context (potentially corrupted) and action reward distributions relevant for action selection in the contextual bandits. The context and transition history can provide the agent with information about the underlying state. Thus, in order for an agent to perform well in our setting, it needs to balance its trust in the two sources of information: first, trust the context that can tell the agent the state of the environment, or second, use the information learned about past state-correlations to determine the state.

The algorithmic approach we propose for the above problem setting is similar to several previous works. Combining algorithms that use different sources of information has been studied more extensively in recent years under the umbrella of \textit{online model selection}. Important work in this area includes the EXP3 and EXP4 algorithms~\citep{Auer2003} that aim to combine multiple expert algorithms for decision-making where each expert may use different sources of information to make decisions. Another major step in optimally combining policies is the seminal paper on corralling a band of bandits, where~\cite{Agarwal16} develop an algorithm called \textit{CORRAL}, providing a better balance between exploration and exploitation when combining \textit{learners} instead of experts.
We concentrate on a well-proven strategy of the ``bandit over bandits approach'', where the master algorithm (typically a bandit itself) chooses between the decisions of base algorithms.

Our main contributions are: (1) we formulate a setting combining crucial aspects of two bandit problems: The correlation of states in time is determined by the Markov chain, and a potentially corrupted context provides information about the underlying state; (2) we propose an algorithm for this setting and a learning mechanism for action-to-action correlation; (3) We provide regret bounds for the problem setting and algorithm. (4) we evaluate our algorithm and several of its variants empirically on both simulated and real-world data. The performance is compared against a set of popular algorithms.

We organize the paper as follows: We discuss the background in section \ref{sec:Backgr}, in particular, our problem setting, bandits, and definitions used throughout the paper. In section \ref{sec:meta} we present the meta-algorithm COMBINE and its upper-confidence-bound (UCB) instantiation COMBINE-UCB. 
We present updated equations for our algorithm and analyze its learning dynamics in section \ref{sec:CalTheo}. Regret bounds of the problem setting and algorithm are presented in section \ref{sec:theoretical}.  
In section \ref{sec:Exper}, we describe our simulation environment as well as the setup of the empirical evaluation of our method on the simulation environment and real-world data. We present and discuss the results in section \ref{sec:Results}. In section \ref{sec:Conclu} we conclude the paper.

\section{Background}\label{sec:Backgr}

In this section, we first informally introduce our problem setting. We are primarily concerned with an environment where observations of varying degrees of ``usefulness'' are available to the agent for decision-making. For example, a context (side information in the form of features) allows the agent to learn the correct context to action reward mapping, improving its decision-making capabilities over time. Our particular setting is motivated by mHealth applications, where users may respond to questionnaires as part of the context, which can be unreliable or conflicting. Users might also exhibit varying engagement and conscientiousness in reporting their state. Thus, timely side information for decision-making may only be available to the agent sometimes.

Besides information from contexts, users in a mHealth setting may also have specific goals, such as losing weight, requiring the progression through states to achieve the desired goal. Each of these states may ask for specific interventions. Knowing about the current state is essential. In the absence of any useful context to estimate the current state of the user, a reasonable approach would utilize transitions between states to guide action exploration more effectively. We assume that the underlying evolving state of the user induces correlation or a sequence of actions to be played by the agent. 

The agent's goal in our setting is to weigh the usefulness of context against the usefulness of state correlations. The agent needs to \textit{learn} their respective value through interacting with the environment. We note that ``usefulness'' is defined relative to the agent's ability, that is, the information source is only as useful as the agent's ability to utilize it for decision-making. For example, a complex nonlinear context-to-action reward mapping might provide no benefit to a linear $CB$; using state correlations instead of the context might prove more useful to the agent.

Thus, the exploration and exploitation trade-off not only involves deciding between using the context or correlations (or some combination). The agent also needs to consider that the information sources' usefulness may change as the agent learns to use them better. The time invested in using one information source more effectively will lead to less experience, and thus less performance, with the other. It is a priori unclear which information source is better than the other or which will ultimately have the lowest cumulative regret over some time frame $T$. Choosing the best approach requires a more subtle trade-off between exploration and exploitation: action selection policies that are optimal from the perspective of one information source are not necessarily optimal when using other sources or a combination of them.

\subsection{Contextual bandit problem and LinUCB}\label{subsec:CB}

Our problem setting and algorithm combine two information sources, from contexts and state transitions. As mentioned before, the context provides the agent with a (potentially noisy) observation of the state. The contextual part exhibits similarities to the contextual bandit problem and the popular LinUCB algorithm, respectively. Therefore we find it instructive to start by introducing both. We follow the definition of the contextual bandit problem from~\cite{Langford2007}. At each time step $t \in \{0,1,\dots,T\}$:

%with slight modifications to include information concerning several users similar to~\cite{Li2010}

\begin{enumerate}
    \item  The environment reveals a d\nobreakdash-dimensional feature vector $\mathbf{x}_{t}\in \rm I\!R^d$.
    \item  The agent chooses an action $a_{t}$ from a set $\mathcal{A}$ of $K$ alternatives according to its policy $\pi:  \mathcal{X} \rightarrow \mathcal{A}$. After playing the action $a_{t}$, the action's reward $r_{t,a_{t}}$ is revealed.
    \item The agent updates its policy using the observations of context $\mathbf{x}_{t}$, action $a_{t}$ and action reward $r_{t,a_{t}}$ to improve action selection in future rounds. 
\end{enumerate}

This formulation of the $CB$ problem does not make any assumptions about the specific relationship between contexts and action rewards. Indeed ~\cite{Langford2007} talk about regret compared to the best hypothesis in some hypothesis class. In our formulation, we concentrate on linear hypotheses, that is, we make the linear realizability assumption. The expected reward of action depends linearly on the context. more formally:

\begin{equation}\label{eq:lra}
\mathop{\mathbb{E}}[r_a | \mathbf{x}] = \mathbf{x}^\top \mathbf{\theta}^*_{a}.
\end{equation}

where $\mathbf{\theta}_a^*$ is some coefficient vector associated with action $a$.

The $CB$ problem formulation presented here does not explicitly include noisy or corrupted contexts. For our setting, we adopt the notion of corrupted context from~\cite{Bouneffouf2020}. The agent receives a corrupted context that does not contain any information to learn the correct context for the action reward mapping. This setting is an extension of the $CB$ problem defined earlier. With probability $p_\nu$, the agent receives a corrupted context. How the context is corrupted is governed by the corruption function $\nu: \mathcal{X} \rightarrow \mathcal{X}$. This function is arbitrary and unknown to the agent and it is not possible to recover or otherwise compensate for it. Further, the agent does not know if the context it received was corrupted, even after receiving the reward.

The context the agent receives at every time step is defined as

\[
    \hat{\mathbf{x}}_t = \left\{\begin{array}{lll}
         \nu(\mathbf{x}_t) &\text{with probability} & p_{\nu}, \\
          \mathbf{x}_t&\text{with probability}& 1-p_{\nu}.
    \end{array}\right.
\]

Contexts and action rewards in this formulation are sampled i.i.d. from a joint distribution $D(\mathbf{x},\mathbf{r})$. This setting assumes that the context contains all information to estimate the context of action reward mapping. In our setting, contexts are sampled i.i.d. from a stationary distribution $D_s(\mathbf{x})$ associated with each state. We specify the relationship between $\mathbf{x}$, $r$ and $s$ more formally in section \ref{sec:ProbSett}.

One popular algorithm is LinUCB~\citep{Chu2011, Li2010}, based on the idea of using UCB methods not only for the $MAB$ problem but also for the $CB$ problem.
The key challenge for UCB-style methods, in general, is the construction of upper confidence bounds on the estimated action rewards and doing so in an efficient manner. This is important since the agent chooses the action with the highest value UCB to manage the exploration/exploitation trade-off. For action rewards that depend linearly on the context,~\cite{Li2010} provides an efficient and closed-form computation rule for constructing confidence intervals.

We focus on LinUCB with disjoint linear models described by~\cite{Li2010}. As in our setting, the linear realizability assumption holds. 
Coefficients $\mathbf{\theta}^*_{a}$ in equation \ref{eq:lra} are unknown to the agent and need to be estimated from collected data. The estimated coefficient vector $\hat{\mathbb{\theta}}_{a}$ can be found via ridge regression:

\begin{equation}
    \hat{\mathbf{\theta}}_{a} = (\mathbf{D}^\top_{a} \mathbf{D}_{a} + k\mathbf{I}_d)^{-1} \mathbf{D}^\top_{a} \mathbf{b}_{a},
\end{equation}

where $\mathbf{D}_{a}$ is a design matrix with dimension $m \times d$ that contains $m$ contexts for action $a$ previously observed, $\mathbf{b}_{a}$ is the reward vector for action $a$ and $\mathbf{I_d}$ is the $d \times d$ identity matrix. $k > 0$ is a parameter we set to $1$. 

At each time step $t$, LinUCB selects actions according to

\begin{equation}\label{eq:LinUCBselect}
    a_{t} = \argmax_{a\in{A}} \bigg(\mathbf{x}^\top_{t}\hat{\theta}_{a} + \alpha\sqrt{\mathbf{x}^\top_{t} \mathbf{A}_{a}^{-1}\mathbf{x}_{t}}\bigg),
\end{equation}

where $\mathbf{A}_{a} = \mathbf{D}^\top_{a} \mathbf{D}_{a} + \mathbf{I}_d$, $\log$ being the natural logarithm and $\alpha = 1 + \sqrt{\log(2/\delta)/2}$, with $\delta > 0$, is a constant determining the level of exploration.  

As a side note, the authors of LinUCB mention that the confidence interval (second term in equation \ref{eq:LinUCBselect}), may also be derived from Bayesian principles using $\hat{\mathbf{\theta}}_a$ and $\mathbf{A}^{-1}_a$ as mean and covariance to the Gaussian posterior distribution of coefficient vector $\hat{\mathbf{\theta}}$. This observation has been utilized in the LinTS algorithm~\citep{agrawal2014thompson}.

\subsection{Regime switching bandits and a discounted UCB algorithm}

The second ingredient to our problem setting includes the non-stationarity of the action rewards and context distribution modulated by an evolving hidden state. This mechanism on the action rewards is similar to \textit{regime switching bandits}~\citep{Zhou2021}. We adopt part of their definition of the problem setting and start out by introducing the setting more formally. In the regime-switching bandit problem, there exists a finite state Markov chain $\mathcal{M}$. This Markov chain evolves according to a transition kernel $\phi(s)$ and each state is associated with a reward distribution depending on the state and the action. At each time step $t \in \{0,1,\dots,T\}$:

\begin{enumerate}
    \item The agent plays action $a_t$ from a set $\mathcal{A}$ of $K$ alternatives according to its policy $\pi:  \mathcal{X} \rightarrow \mathcal{A}$.
    \item The action's reward $r_{s_t, a_t}$ according to the reward distribution $R(\cdot|s, a) := P(r_t \in \cdot|s = s_t, a = a_t)$ is revealed.
    \item the agent updates its policy using reward $r_{s_t, a_t}$
    \item the environment advances the Markov chain $\mathcal{M}$ by sampling the next state according to the transition kernel $\phi(s)$. 
\end{enumerate}

The agent has \textit{no} a priori knowledge of either $\phi$, the state $s$, or reward distribution $R$ and needs to learn about these quantities from collected data. 

Previous settings with this particular property are \textit{rested} bandits~\citep{Gittins1979}, where the reward of each action is coupled to a separate finite state Markov chain. Only the state of the chosen action evolves. The states of other actions remain ``frozen''. A more general setting is the \textit{restless} bandits~\citep{Whittle88, Ortner2014}, where states of other actions may also evolve at each time step. A common Markov chain that determines the reward distribution of \textit{all} actions is what separates regime-switching bandits from these previously investigated settings. 

Agents in the regime-switching bandits setting cannot access side information commonly available in practical applications. Side information allows partial observation of the hidden state. This differentiates our setting from theirs, where we allow a noisy observation coupled with the hidden state to be used for decision-making, allowing the agent to differentiate states by context.

The authors in the regime-switching bandits paper provide a computationally inefficient algorithm based on spectral estimation, solving for an optimistic plausible POMDP. To aid the interpretation and analysis of the experimental results, we opt for simpler and more efficient algorithms that are augmented with the ability to utilize gained knowledge about state transitions. As mentioned earlier in the introduction, several state-of-the-art algorithms deal with non-stationary action reward distributions. One of the earliest and most straightforward approaches involves using algorithms developed for the stationary $MAB$ problem and equipping them with the ability to \textit{discount} prior experience about action rewards. We introduce a discounted version of the UCB algorithm as an illustrative example since we also use an adapted version in our algorithm. The idea behind this modified algorithm is not much different from the classical UCB algorithm~\citep{Sutton2018}, that is, the agent chooses the action with the highest \textit{upper confidence bound} (UCB) on the mean rewards collected so far as

\begin{equation}
    a_t = \argmax_{a \in \mathcal{A}}\; UCB(a).
\end{equation}

The UCB for each action is computed as

\begin{equation}\label{eq:ducb}
    UCB(a) = \bar{R}_t(a) + \alpha_B\sqrt{\frac{\log(t)}{N_t(a)}}, 
\end{equation}

where $\log$ is the natural logarithm, $\bar{R}(a)$ is the mean reward of action $a$ and $N_t(a)$ is the number of times action $a$ has been played so far. $\alpha_{B} \in (0,\infty]$ is a scalar that modulates the level of exploration. The discounted variant we use computes $\bar{R}(a)$ as a discounted sum. Let $r_n(a)$ be the reward received at the $n^{th}$ play of action $a$, then 
\begin{equation}\label{eq:avg_reward_sum}
    \bar{R}_{N_t(a)+1}(a) = \bar{R}_{N_t(a)} + \gamma(r_{N_t(a)} - \bar{R}_{N_t(a)}) = \sum_{n=1}^{N_t(a)} \gamma(1-\gamma)^{{N_t(a)}-n} r_{n}(a),
\end{equation}

assuming $\bar{R}_0(a) = 0$. The scalar $\gamma\in(0,1]$ determines the weight of recent rewards on the mean $\bar{R}(a)$. This simple modification allows the agent to adapt more quickly in the non-stationary setting when the distribution of action rewards changes.

Discounted UCB (D-UCB) is an alternative that additionally discounts the number of action plays. We do not discount the number of action plays in our implementation. We come back to D-UCB in the theoretical analysis in section \ref{sec:theoretical}.

While a straightforward and convenient method, prior knowledge about the action rewards is inevitably discarded (or discounted). Referring back to the regime-switching setting, if the environment revisits states, the agent needs to \textit{relearn} the action rewards, thus needlessly suffering regret. We introduce a modification later in section \ref{sec:meta} that helps in mitigating this shortcoming. We equip the agent with the ability to focus on a reduced subset of actions that show correlations in rewards between states. This is achieved by learning an action-to-action adjacency matrix. Details of the exact procedure to learn the matrix are presented in section \ref{sec:updateadjmatrix}.  

\subsection{Formal problem setting}\label{sec:ProbSett}

In this section, we describe the problem setting, introduced previously in an intuitive way, using formal notation. We consider an online learning problem with bandit feedback, where contexts can be corrupted, and the context and action reward distributions are determined by an underlying hidden state evolving according to a Markov chain. There exists a set of users $I$ constituting separate environments. Thus, while practically relevant, the existence of $I$ is not required for our formal problem definition. The hidden states come from a finite set $\mathcal{S}$ with cardinality $|\mathcal{S}| = S$, and actions come from a finite set $\mathcal{A}$ with cardinality $|\mathcal{A}| = K$. Each state is associated with a stationary context distribution $D_s$, for example, Gaussian $\mathcal{N}(\mathbf{\mu_s}, \Sigma_s)$ with mean $\mathbf{\mu_s} \in \rm I\!R^d$ and covariance $\Sigma_s \in \rm I\!R^{d\times d}$. Contexts are sampled i.i.d. from $D_s$. Further, we assume the expected reward is a linear function of the context, that is, $\EX[r_{a,s} | \mathbf{x}_t] = \mathbf{x}_t^{\top} \theta_a(s) = \mu_a(s, \mathbf{x}_t)$, where $\theta_a(s)$ is an unknown weight vector associated with action $a$ in state $s$. $\mu_a$ is the mean reward for action $a$ in state $s$ and context $\mathbf{x}_t$. $\theta_a(s)$ needs to be learned from data. We define the optimal arm at time $t$ as $a^*_t = \argmax_{a \in \mathcal{A}} \mu_a(s_t, \mathbf{x_t})$. Denote $\pi:  \mathcal{X} \rightarrow \mathcal{A}$ as the policy. We focus on Bernoulli rewards since, in practice, rewards are often collected in the form of "like/dislike" or "click/no click" on an advertisement (however, extension to arbitrary rewards is quite natural). Two sources of information are available to the agent: corrupted contexts and state correlations that result in time correlation among actions. Learning these correlations can help reduce the set of actions to consider, depending on the belief about the current state. The agent needs to decide which source of information to trust more for decision-making. We model this problem via a repeated game between the agent and the environment (user). At every decision point $t=1,2, \dots, T:$
\begin{enumerate}
    \item The environment draws a context $\mathbf{x}_{t}$ from $D_s$ for current state $s_t$;
    \item The context is corrupted $\hat{\mathbf{x}}_{t} = \nu(\mathbf{x}_{t}),\nu: \mathcal{X} \rightarrow \mathcal{X}$ with probability $p_{\nu}$;
    \item The possibly corrupted context $\hat{\mathbf{x}}_{t}$ is revealed to the agent;
    \item The agent chooses an action $a_{t} = \pi(\hat{\mathbf{x}}_{t})$;
    \item The environment reveals the reward $r_{a_{t}, s_{t}} \in \{0, 1\}$;
    \item The state $s_t$ is updated: $s_{t+1} = \phi(s_{t})$,
\end{enumerate}

where $\phi: \mathcal{S}\rightarrow \mathcal{S}$ denotes a function determining the next underlying state. We also demonstrate, through experiments on real-world data sets in section~\ref{sec:datasets}, that it is a realistic assumption. In general, for finite state spaces, $\phi$ function takes the form of a transition kernel. Thus, the state $s_{t}$, is not drawn independently from an underlying stationary distribution, like in~\cite{Bouneffouf2020}, but evolves according to a finite-state Markov chain.

The difference in cumulative reward between the policy that always chooses the optimal action on every time step and the learned policy $\pi$ of the agent is commonly referred to as the \textit{dynamic regret}. We compute the regret in the following.

\paragraph{Dynamic regret} The dynamic regret after $T$ time steps defined as

\begin{equation}
    R(T) = \sum^T_{t=1} \mu_{a_t^*} - \mu_{a_t},
\end{equation}

where $a^*_{t}$ and $a_{t}:= \pi(\mathbf{\hat{x}}_{t})$ denote the optimal action and the action chosen by the agent's policy at time step $t$, respectively. When defining regret in this way, we note that our agent competes against the oracle that chooses the best action at every time step. Given that the hidden state evolution is Markov, an algorithm can perform much better by playing the best action in each state, compared to playing the action with the highest mean reward \textit{over all states and time steps}, that is, the highest reward action in the action set $\mathcal{A}$~\citep{Ortner2014}. 

\section{Meta-Algorithm: Competing Bandits with Corrupted Context and Action Correlations}\label{sec:meta}
\setcounter{algorithm}{0}
\begin{algorithm}
\caption{Competing Bandits with Corrupted Context and Action Correlations}
\label{alg:alg2}
\begin{algorithmic}[1]

\Procedure{COMBINE}{} 
    \State \textbf{Input:} Algorithm parameters, Policies: $CB$, $MAB$ and referee, action set, user set
    \State \textbf{Initialize:} Book-keeping variables for $CB$, $MAB$ and referee
    \For{$t=1,2,\dots,T$}
        \For{user $i\in I$}
            \State Observe context $\mathbf{\hat{x}}_{i,t}$
            \State Sample policy $\pi(t)$ from referee
            \If{$\pi_{i,t}$ = $CB$ policy} 
                \State choose action using $CB$ policy
            \Else 
                \State choose action using $MAB$ policy from subset ${\mathcal {U}_i}$
            \EndIf
            \State Observe the reward for the chosen action
            \If{previous indicator action $\neq$ the current chosen action}
                \State Update adjacency matrix $\Lambda^i$
            \EndIf
            
            \If{CB was chosen as a policy}
                \State Update $CB$ policy
            \Else 
                \State Update $MAB$ policy
            \EndIf
           
            \State Update referee
            
            \If{reward = 1} 
                \State Update current indicator action $a^+_i$
            \Else
                \State Update previous indicator action $a^-_i$
            \EndIf
            \State choose action subset ${\mathcal {U}_i}$ to sample from next 
        \EndFor
    \EndFor
\EndProcedure

\end{algorithmic}
\end{algorithm}

We present the high-level overview of the meta-algorithm \textit{COMpeting BandIts with corrupted coNtext and action corrElations} (COMBINE) in algorithm~\ref{alg:alg2}, and its UCB variant, in algorithm~\ref{alg:alg2.1}. The key idea is that at each time step, after the agent observes the context $\mathbf{\hat{x}}_t$ (possibly corrupted), it decides to use either the $CB$ or $MAB$ policy (line 7). The $MAB$ policy, in particular, chooses the actions to play from a reduced subset of actions determined by the learned adjacency matrix $\Lambda^i$. Therefore, we call the $MAB$ policy $MAB_u$. 
The decision to use the $CB$ or $MAB_u$ is made based on the expected reward of each policy, estimated from the past performance by another bandit, the so-called referee.

If the referee chooses the $MAB_u$ policy, the $MAB_u$ is presented with the actions from an action-subset $\mathcal {U}$ to choose from. $\mathcal {U}$ is dynamically computed and represents a candidate set of next promising actions. This candidate set is computed using the $\Lambda^i$, which counts the number of transitions observed between actions that provide a reward of $1$. If the $CB$ is chosen, actions are selected according to the context from the complete action set. 
The selected action is played, and the agent observes the reward. 

If the $CB$ policy was chosen, it is updated with the received reward (line 15). Otherwise, we update the $MAB_u$ policy (line 18). Our reason here is that the performance of the $CB$ is worse compared to the $MAB_u$. The $CB$ may not gain anything by learning from the data point. Discarding seemingly corrupted contexts will lead to more efficient learning of context-to-action reward mapping. Similarly, if the $MAB_u$ has difficulty selecting the appropriate action, it may only harm its estimates of true action rewards when the state changes frequently. The referee is then updated using the reward and the choice of policy.

As mentioned before, the algorithm keeps track of state transitions through an action-to-action adjacency matrix $\Lambda^i$ for each user $i$ that contains the number of observed transitions between the algorithm's chosen actions. The matrix is updated after the reward is observed, provided that a potential transition into another state has occurred (line 13). To determine a change, we rely on the fact that most of the time, the algorithm will choose the action that provides the highest reward in the state. If the action selection changes, the adjacency matrix records a transition between the previously chosen action and the currently chosen action.

As part of the adjacency matrix update, we store and update two \textit{indicator actions}, that is, actions for which the agent received a reward of $1$ (Bernoulli distributed rewards). These two indicators, \textit{previous indication action} $a^-_i$ and \textit{current indicator action} $a^+_i$, are used to record what action leads to another.  The $MAB_u$ uses the adjacency matrix $\Lambda^i$ to select the next promising set of actions. For example, if a change in rewards is detected, the choice is limited to the probable next "best" actions to a subset $\mathcal{U}_i$, that is, $\mathcal{U}_i = \Lambda^i(a^-, \cdot) \geq 1$. Finally, the action subset $\mathcal{U}_i$ is updated, which is used in the next round (line 24).

\subsection{Competing Bandits with Corrupted Context and Action Correlations UCB (COMBINE-UCB)}

\begin{algorithm}
\caption{Competing Bandits with Corrupted Context and Action Correlations UCB (COMBINE-UCB)}
\label{alg:alg2.1}
\begin{algorithmic}[1]
    \State \textbf{Input:} $\alpha \in (0,\infty]$, $\alpha_B \in (0, 1]$, $\gamma \in (0,\infty]$, $\delta_R \in (0,\infty]$, action set $A$, user set $I$
    \State \textbf{Initialize:} preference values $H=\{0\}^{|I|\times2}$, average action reward $\Bar{R} = \{0\}^{|I|\times|A|} $, action play count $n=\{0\}^{|I|\times|A|}$, current indicator action $a^+ = \{NaN\}^{|I|\times1}$, previous indicator action $a^- = \{NaN\}^{|I|\times1}$, reach parameter $\beta = \{1\}^{|I|\times1}$, action subset ${\mathcal {U}}^{|I|\times |A|} = \mathcal{A}$, adjacency matrix per user $\Lambda^i = \{1\}^{|A|\times|A|}$, design matrix of LinUCB $\forall a\in \mathcal{A} : \mathbf{A}_a = \mathbf{I_d}$, reward vector $\forall a\in \mathcal{A} :\mathbf{b_a} = \{0\}^{d\times1}$, upper-confidence action score $UCB = \{0\}^{|I|\times K}$
    \For{$t=1,2,\dots,T$}
        \For{user $i \in I$}
            \State Observe context $\mathbf{\hat{x}}_{i}$
            \State $pb = \frac{e^{H^i}}{\sum e^{H^j}}$ \Comment{probability of choosing either $CB$ or $MAB_u$}
            \State Sample policy $\pi^i \sim {\mathcal {B}}\Big(pb^i\Big)$
            \If{$\pi^i = \pi_0$} \Comment{Choose global $CB$}
                \For{action $a = 1, \dots, K$}
                    \State $\hat{\theta}_a \gets \mathbf{A}_a^{-1} \mathbf{b}_a$ \Comment{Update weight vectors of global $CB$}
                \EndFor
                \State $a^i = \argmax_{a\in{1,\dots,K}} \hat{\mathbf{x}}_{i}^\top\hat{\theta}_a + \alpha\sqrt{\hat{\mathbf{x}}_{i}^\top \mathbf{A}_a^{-1}\hat{\mathbf{x}}_{i}}$ 
            \Else \Comment{Choose $MAB_u$}
                \State $a^i = \argmax_{a\in {\mathcal {U}}_i} UCB_{a,i}$ 
            \EndIf
            \State Observe reward $r^i_{a^i_t, s^i_t}$ for action $a^i$
            \If{$a^-_i \neq a^i$} \Comment{Update adjacency matrix}
               \State $\Lambda^i_{a^-_i,a^i} = \Lambda^i_{a^-_i, a^i} + r^i_{a^i_t, s^i_t}$ 
               %\State $D_{k(t),k^+} = D_{k(t),k^-} + r_{k(t)}$ \Comment{Symmetric update}
            \EndIf
            
            \If{$\pi^i = \pi_0$} \Comment{Update global $CB$ if it was chosen as a policy}
                \State $\mathbf{A}_a\gets \mathbf{A}_a + \hat{\mathbf{x}}_{i}\hat{\mathbf{x}}_{i}^\top$
                \State $\mathbf{b}_a \gets \mathbf{b}_a + r^i_{a^i_t, s^i_t}\hat{\mathbf{x}}_{i}$
                \algstore{combineucb}
\end{algorithmic}
\end{algorithm}

\begin{algorithm}
\begin{algorithmic}
            \algrestore{combineucb}
            \Else
                \State $\Bar{R}^i_{a^i}= \Bar{R}^i_{a^i} + \gamma \big(r^i_{a^i_t, s^i_t}-\Bar{R}^i_{a^i}\big)$ \Comment{Update average action reward}
                \State $n_{a^i,i} = n_{a^i, i} + 1$  \Comment{Update action count for chosen action}
                    \State $UCB_{a^i,i}\gets \Bar{R}^i_{a^i} + \alpha_B\sqrt{\frac{2ln(t)}{n_{a^i, i}}}$ \Comment{Update UCBs}
            \EndIf

            \State $H^i_{\pi^i} \gets H^i_{\pi^i} + \delta_R\big(r^i_{a^i_t, s^i_t}-pb^i_{\pi^i}\big)$  \Comment{Update gradient bandit for $\pi^i$}
            \State $H^i_{\neg \pi^i} \gets H^i_{\neg \pi^i} + \delta_R \big(1 - 2r^i_{a^i_t, s^i_t}\big) \big(1 - pb^i_{\pi^i}\big)$ \Comment{Update gradient bandit for $\pi \neq \pi^i$}
            \If{$r^i_{a^i_t, s^i_t} = 1$} 
                \State $a^+_i \gets {a^i}$ \Comment{Update current indicator action}
            \Else
                \State $a^-_i \gets a^+_i$ \Comment{Update previous indicator action}
            \EndIf
            ${\mathcal {U}}_{i} \gets AdjSelect(a^-_i, \Lambda^i, \beta_i, r^i_{a^i_t, s^i_t})$ \Comment{Update action subset}
        \EndFor
    \EndFor
\end{algorithmic}
\end{algorithm}

\begin{algorithm}
\caption{Select Action Subset (AdjSelect)}
\label{alg:alg3}
\begin{algorithmic}[1]

    \State \textbf{Input:} previous indicator action $a_i^-$, adjacency matrix $\Lambda^i$, reach $\beta_i$, reward $r^i_{a^i_t, s^i_t}$
    
    \If{$\forall a: UCB_{a, i} = 0$ or $a_i^- = NaN$} \Comment{Never have played an action before, or no previous indicator action}
        \State \textbf{return} $\mathcal{A}$ \Comment{Return complete actions set}
    \Else
        \If{$r^i_{a^i_t, s^i_t} = 1 $} 
            \State $\beta_i \gets 0$ \Comment{Reset reach}
        \Else
            \State $\beta_i \gets min\{\beta_i + 1, K\}$ \Comment{Increase reach}
        \EndIf
        \State Sort row entries of adjacency matrix $\Lambda^i_{a^-}$ in descending order, creating set $S^i_{a^-}$
        \State Select top $\beta_i$ entries in $S^i_{a^-,\{1,\dots,1+\beta\}} := \mathcal{U}_i$ 
        \State \textbf{return} ${\mathcal {U}}_i$
    \EndIf
\end{algorithmic}
\end{algorithm}

The UCB instantiation of the meta-algorithm presented in the previous section is shown in algorithm \ref{alg:alg2.1}. At every time step, the agent observes a (possibly corrupted) version of the context $\mathbf{\hat{x}}_{i}$. The decision to use the $CB$ or $MAB_u$ policy to select an action is performed by a gradient bandit ~\citep{Sutton2018}. The probability of choosing $CB$ or $MAB_u$ is computed from the preference values $H_i$ (line 6) using the softmax distribution. The policy is sampled from a Bernoulli distribution with parameter $p = pb_i$. Suppose the referee picks the contextual bandit (policy $\pi_0$). In that case, using a global model, the agent chooses the action that maximizes the expected reward plus an upper confidence term like in standard LinUCB~\citep{Li2010} (line 9). Otherwise, the agent selects the action based on the particular user's $MAB_{u}$ strategy using equation \ref{eq:ducb} (line 12). Before choosing the action with the highest upper-confidence score ($UCB_{a, i}$), the subset of potential action candidates for exploration is computed by algorithm \ref{alg:alg3} using the individual action-to-action adjacency matrix for every user. After playing the action and observing reward $r^i_{a^i_t, s^i_t}$, we update the parameters of the policy chosen by the referee (line 18). Finally, depending on the received reward, we update the preference values $H^i$ (line 25). For example, if the $CB$ policy was chosen, that is, $\pi^i = \pi_0$ and a reward $r^i_{a^i_t, s^i_t} = 1$ was received, we increase the preference $H^i_{\pi^i_{t}}$  for choosing the $CB$ while decreasing the preference for the $MAB_u$ policy $H^i_{\neg \pi^i}$.

Algorithm \ref{alg:alg3} is a simple strategy to dynamically adapt the subset of actions to play in each round. Using \textit{reach} $\beta$, the algorithm selects the top $\beta$ entries of the $a_i^-th$ row of individual adjacency matrix $\Lambda^i$, sorted in descending order. The reach parameter can be thought of as the size of the set of actions to play. The larger the reach, the more actions are included. Actions that show correlations in a first-order Markov chain are treated with higher priority and are included first. The algorithm greedily chooses the action subset (see lines 6-8) depending on recent rewards. It can adapt quickly to changing reward distributions by increasing $\beta$ rapidly, thus including actions with less strong correlation to ensure sufficient exploration by the action selection algorithms. Note that algorithm \ref{alg:alg3} does not compute an explicit term (like UCB-type algorithms) to ensure asymptotic optimally by enforcing optimistic behavior. Nonetheless, sufficient exploration is ensured simply by the greediness of the algorithm when selecting the pool of possible candidate actions. When the chosen action does not provide a positive reward (rewards are Bernoulli distributed), other actions are included for potential exploration by the $MAB_u$ strategy.

\section{Calculations and Theory}\label{sec:CalTheo}

In this section, we detail the calculation steps for different parts of the presented algorithm. Furthermore, we discuss the implications of some of the design choices on the algorithm's behavior.

\subsection{Update equations of the gradient bandit}
The probability $pb^i$ of choosing the $CB$ or the $MAB_u$ is computed from a gradient bandit's~\citep{Sutton2018} preference values. The preference values can be interpreted in the following manner: the less reward one of the base algorithms receives, the less likely the gradient bandit is to choose the base algorithm. If the $MAB_u$ performs worse than the $CB$, the preference values for the $MAB_u$ become \textit{smaller} compared to the $CB$. For example, after some time, the agent may have for $H^i =\{1, -2\}$. The probability of choosing the $CB$ (policy $\pi_0$) is then $ pb^i_{\pi_0} = \frac{e^{1}}{e^{1}+ e^{-2}} \sim 95\%$.
The updates are

\begin{align*}
H^i_{\pi^i} &\gets H^i_{\pi^i} + \delta_R\big(r^i_{a^i_t, s^i_t}-pb^i_{\pi^i}\big),\\
H^i_{\neg \pi^i} &\gets H^i_{\neg \pi^i} + \delta_R \big(1 - 2r^i_{a^i_t, s^i_t}\big) \big(1 - pb^i_{\pi^i}\big).
\end{align*}

We modified the update equation of the gradient bandit presented in~\cite{Sutton2018}, since the factor $(1- \pi^i)$ in the original definition of $H^i_{\pi^i}$ may favor algorithms with strong initial performance in the update, thus reducing adaptability in the non-stationary environment due to vanishing gradients when probabilities are close to $1$~\citep{Mei2020}. 

We also omitted the comparison of the received reward $r^i_{a^i_t, s^i_t}$ and average reward $\Bar{R^i_t}$ as in the original definition. This comparison served as a baseline to modulate the magnitude and direction of the preference value update (second term on the RHS in equations above). Omitting this comparison allows for faster adaptability of the preference values since $\Bar{R^i_t}$ generally reduces the magnitude of the update. Our formulation would punish policies more strongly for mistakes if they were chosen more often. The gradient bandit can adapt much more quickly to changes in reward performance of the action selection policies, albeit with the caveat of being potentially more susceptible to reward noise. Note that the reward the gradient bandit receives depends on the performance of the chosen policy. Thus, the referee and chosen bandit are updated in a joined manner via the referee's choice of base algorithm and the base algorithm's action selection.

\subsubsection{Gradient bandit dynamics}

The learning dynamics of the gradient bandit can be described by a solution in the following nonlinear differential equation (see appendix \ref{sec:ap} for the derivation)

\begin{equation}\label{eq:diff}
\frac{d p^*(t)}{d t} = \delta_R
p^*(t)(\Delta_R - r^* + p^*(t)(2 p^*(t) - 3)(\Delta_R p^*(t) - r^* + 1) + 1),
\end{equation}

\noindent where $r^*$ is the average reward of the superior policy, $p^*(t)$ is the probability of choosing the superior policy at time $t$, and $\Delta_R$ is the gap in average reward between the two policies. $p^*(t)$ reaches the following stationary probability of choosing the better-performing policy:
\begin{equation}\label{eq:limit}
\lim_{t\to\infty} p^*(t)= \min \Bigg\{1,\; \frac{\Delta_R + 2r^* + \sqrt{9\Delta_R^2 - 4\Delta_R r^* + 4\Delta_R + 4 r^{*2} - 8r^* + 4} - 2}{4\Delta_R}\Bigg\} \\ := C_\infty.
\end{equation}
Equation \ref{eq:limit} follows readily from setting the LHS of equation \ref{eq:diff} to $0$ and solving for $p^*(t)$. Note that we employed the $\min$-operator, since \ref{eq:diff} may have roots outside the domain $[0, 1]$.  

\begin{figure}
    \centering
    \includegraphics[width=.8\linewidth]{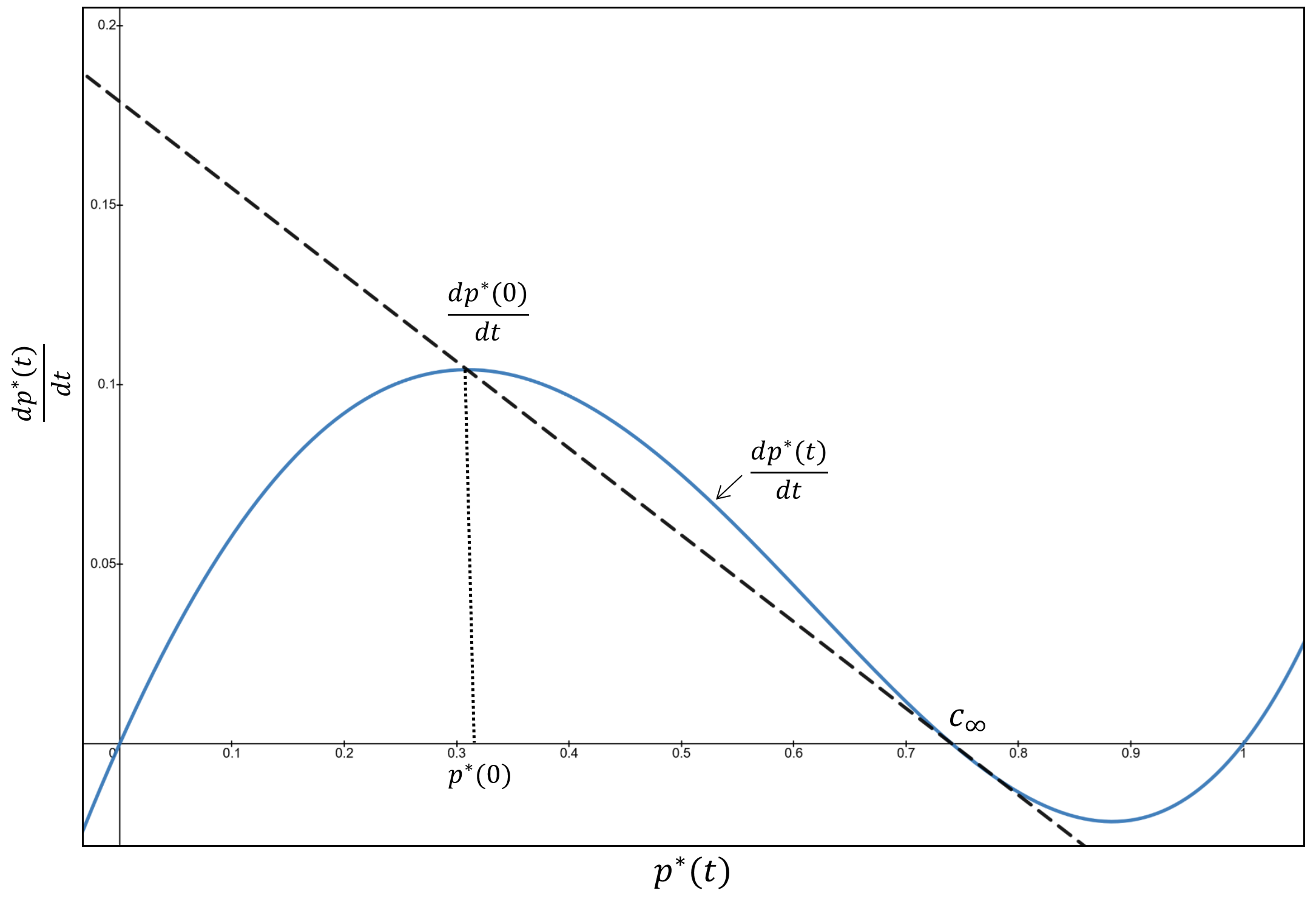}
    \caption{$dp^*(t)/dt$ vs. $p^*(t)$ for $\delta_R = 1$, $\Delta_R = 0.3$ and $r^* = 0.6$. The secant line provides a lower bound on $dp^*(t)/dt$, for $p^*(t) < C_{\infty}$. The derivative becomes linear. Thus, we have an exponentially decreasing function as the solution $p^*(t) \propto e^{-t}$.}
    \label{fig:secant_line}
\end{figure}

We do not solve for an explicit solution of $p^*(t)$, but for a constant $\Delta_R$ we can approximate it rather well by a simple exponential function, by defining the $p^*(t)$ as the secant line between $p^*(0)$ and $C_{\infty}$, thus being a lower bound on $dp^*(t)/dt$(see figure \ref{fig:secant_line} as an illustration). 

The convergences goes with approximately $e^{-t\Delta_Rc}$, where $c>0$ is some constant. Using equation \ref{eq:limit}, we let  $p^*(t)$ be (probability of choosing the superior policy, starting from $p^*(0)$ and ending at $C_{\infty}$)

\[p^*(t) \approx \big(C_\infty-p^*(0)\big)(1- e^{-t\Delta_Rc})+p^*(0),\]

with $\Delta_R > 0$ and some problem-dependent constant $c\neq0$.   We integrate over $T$ to get the number of times the superior policy is chosen, that is

\begin{multline}
    \tau^* \sim \int_0^T \big(C_\infty-p^*(0)\big)(1- e^{-t\Delta_Rc})+p^*(0) dt=\\
\frac{1}{c\Delta_R} \big[ e^{-T\Delta_Rc} \big(C_\infty(1-e^{T\Delta_Rc})+ p^*(0)( e^{T\Delta_Rc}- 1) \big) \big] + C_\infty T \label{eq:equilib}.
\end{multline}
Therefore, the number of time steps the inferior policy is chosen is sublinear in $T$ (first term in equation \ref{eq:equilib}) when $C_\infty$ (second term) converges to $1$ eventually. From this analysis, we see that our gradient bandit does not commit as readily to the best policy compared to the vanilla formulation by~\cite{Sutton2018}. Instead, it is preserving fast adaptability in a changing environment, with the caveat that additional regret may be incurred in scenarios where $\Delta_R$ is small, and consequently $C_\infty < 1$. This behavior of our gradient bandit can be readily observed in the section \ref{sec:Results} and leads to a dynamic equilibrium between choosing the $CB$ and $MAB_u$ policies according to the performance relative to each other.

\subsection{Updating the Adjacency Matrix of COMBINE}\label{sec:updateadjmatrix}

We assume that we receive information from several users at each time step. For COMBINE-UCB and variants, we assume that users in a state share the same context distribution. Thus the learned action to reward mapping from each user could be generalized to other users. The meta-algorithm allows the possibility of including an individual or global model for the $CB$, $MAB_u$, and referee. We choose a global model for the $CB$ and individual models for both $MAB_u$ and the referee. Both our $MAB_u$ policy and the referee can adjust to each user.
 We also investigate learning a common adjacency matrix for all users. The individual $MAB_u$ policies use this common adjacency matrix for decision-making. We investigate four ways of updating and utilizing the adjacency matrix for use in the multi-armed bandit part of COMBINE, described in the following subsections.

\subsubsection{Adjacency Matrix}
Each user has an adjacency matrix, which the agent uses to select promising subsequent actions. The two bookkeeping variables, the previous indicator action denoted as $a^-_i$ and the current indicator action denoted as $a^+_i$, are used to record the time-correlation of actions. If the previous indicator action is different from the current action, the algorithm did not receive a reward in the previous step (a transition occurred potentially) and may need to explore. We then update the adjacency matrix according to
 
\begin{equation}
        \Lambda^i_{a^-_i, a^i} = \Lambda^i_{a^-_i,a^i} + r^i_{a^i_t, s^i_t}.
\end{equation}

If the agent received a reward $=1$ for the currently played action $a_{i}$, we set $a^+_i \gets a^i$, otherwise, we set $a^-_i \gets a^+_i$.

While we cannot be sure that using this mechanism recovers the actual correlation among actions, it does so in expectation. As the agent learns the best action, it selects it more frequently and sooner. Errors corresponding to lower rewards give the algorithm an indication that something has changed. However, there is no mechanism to guarantee this, given that the agent may always make mistakes in estimating the current state. If something more is known about the setting, for example, about the frequency of state changes, more elaborate solutions might be possible.

There might be cases where users behave similarly over time, where we can learn the adjacency matrix jointly by combining the experience gained from multiple users. We investigate a variant where each user contributes to a common adjacency matrix. Referring back to the meta-algorithm, adjacency matrix $\Lambda^i$ (line 14 in algorithm \ref{alg:alg2}) would constitute a global model of the action transitions. The common adjacency matrix $\Lambda_{com}$ is computed as the sum over all adjacency matrices at time $t$ for users in $I$ as 

\begin{equation}
    \Lambda_{com,t} = \sum^{|I|}_{j=1} \Lambda^j_t.
\end{equation}

\subsubsection{Softmax Variation}
We investigate a variant where we model transition probabilities for each action pair as the softmax distribution of a vector of action preferences. The adjacency matrix $\Lambda^i$ becomes an action preference matrix $\Lambda^*_i$ and is updated iteratively according to

\begin{equation}
    \Lambda^{*,i}_{a_i^-, a^i} = \Lambda^{*,i}_{a_i^-, a^i} + \alpha_S \Bigg(r^i_{a^i_t, s^i_t}- \frac{e^{\Lambda^{*,i}_{a_i^-,a^i}}}{\sum^{|\mathcal{A}|}_{b=1} e^{\Lambda^{*,i}_{a_i^-,b}}}\Bigg),
\end{equation}

where $\alpha_S \in (0,\infty]$ is a step size parameter that modulates the rate of change of the preference matrix between time steps. The higher the value of this parameter, the larger the update towards the new preference matrix. Instead of simply adding the number of occurrences of some arbitrary transition  $a^i_{t} \rightarrow a^i_{t+1}$ with $a^i_{t} \neq a^i_{t+1}$, particular transitions given the previous indicator action are more or less ``preferred'' by the agent. Subsequently, line 14 in algorithm \ref{alg:alg2.1} changes from the $\argmax$ over actions to sampling from a probability distribution over the action preferences. The probability of choosing action $a^i$ at time $t$ then becomes

\begin{equation*}
    P(a^i|a_i^-) = \frac{e^{\Lambda^{*\beta_i,i}_{a_i^-,a^i}}}{\sum^{|\mathcal{A}|}_{b=1} e^{\Lambda^{*,\beta_i,i}_{a^-,b}}},
\end{equation*}

where the column-indices (next possible actions) of $\Lambda^*_i$ are limited by reach $\beta$ and can vary from time step to time step, resulting in action preference vector $\Lambda^{*,\beta_i,i}_{a_i^-}$.   

As with the common adjacency matrix described above, we define a common preference matrix $\Lambda^*_{com}$, which is computed as a weighted sum of individual preference matrices using the hamming product as
    \begin{equation}
        \Lambda^*_{com,t} = \sum^{|I|}_{j=1} \Lambda^{*,j}_t\circ \Lambda^j_t \oslash \Lambda_{com,t},
    \end{equation}
    
where $\oslash$ denotes the element-wise division. The computational rule can be explained intuitively: the more we observe a particular transition from a user, the more we are sure about said transition for all users, assuming that users behave similarly. We encode this fact by using the adjacency matrix $\Lambda^j_t$, which puts more weight on collected data from a particular user where the agent has observed a specific transition more often.

\section{Theoretical Analysis}\label{sec:theoretical}

This section analyses the problem setting and the algorithm in terms of regret lower and upper bounds, respectively. Our problem setting exhibits both CB and MAB properties, context corruption, and action correlated in time due to the Markovian transitions of the underlying (or hidden) state. Both bounds depend highly on the dynamics of this hidden state in combination with the corruption level. Therefore, we analyze the setting by first considering each context corruption and action correlation individually before providing arguments on regret when these two components interact.

\subsection{Regret lower bound of the contextual bandit problem with corrupted context and action correlations (CBCCAC)}

We generally expect the number of state changes to be of the order less than $T$, justified by the domain-specific phenomenon that people recurrently traverse different stages of behavior change, spending longer and longer time in the "Maintenance" or "Terminal" stage~\citep{prochaska2015}. This assumption is essential for context-free algorithms to achieve sublinear regret. It is particularly relevant in our setting with corrupted contexts, where we may need to rely on context-free algorithms to make decisions. 

In essence, our problem setting exhibits both properties of the contextual bandit problem and restless Markov bandit problem. We expect either the contextual bandit lower bound or switching Markov bandit lower bound holds. We first analyze the contextual and context-free settings and combine them later in our final result.

We begin with the restless Markov bandit. While our problem setting provides information about the current state via the context, corruption of said context will force agents to make decisions without precise knowledge about the current state, thus needing to detect changes in the reward distribution and adjust the policy. In order for an algorithm to detect whether the empirical mean reward of the action $a$ has changed by the amount $\epsilon > \Delta$, with $\Delta$ being the difference in mean reward between arm $a$ and the current best arm, there have to be at least, up to logarithmic factors,  $1/\epsilon^2$ samples of the arm $a$~\citep{Auer19a}. Regarding our problem setting, since all arms are correlated in time, an algorithm can use the change in mean reward of the current best arm to detect a state change. 

The scenario above is reminiscent of the "consecutive sampling" strategy in~\cite{Auer19a}, but it is built into our problem setting. Their results show that depending on the number of states changes $L$ up to time $T$ and the number of actions $K$, a non-trivial lower regret bound can be achieved $O\big(\sqrt{KLT}\big)$, thus lower bounding the regret for the setting where a state change may come with an arbitrary change in reward distribution of actions. This arbitrary change in action reward distribution forces an algorithm to \textit{relearn} the action reward after each switch. In contrast, in our problem setting, the number of states is limited, and they come with a fixed action reward distribution each. Thus, our setting is similar to restless Markov bandits introduced in~\cite{Ortner2014}, and the lower regret bound applies to our problem setting with some modifications. 

\begin{theorem}\label{theo:b}
For any context-free algorithm, the regret after $T$ time steps in the CBCCAC setting, for any $K > 1$, minimum sub-optimally gap $\Delta_{min}$ for all actions, number of state changes L, number of states $S$ and minimum detectable change $\epsilon_{min} > \Delta_{min}$ in mean reward, is lower bounded by 
\[
E[R(T)] \geq \mathcal{O}\bigg(\sqrt{SKT} + \frac{\Delta_{min}\sqrt{KT\Tilde{L}}}{\epsilon_{min}}\bigg),
\]
with

\[
E[\Tilde{L}] \leq L.
\]

\end{theorem}

\begin{proof}
Similar to the argument of~\cite{Ortner2014}, the agent faces $S$ distinct learning problems. The learner must infer the mean rewards for all actions in each state. Imagine a scenario where the learner goes through all states and stays for $T/S$ time steps in each of them. In this case, each problem can be made to force a regret of order $\Omega\big(\sqrt{KT/S}\big)$ in the $T/S$ steps the learner faces a $MAB$ problem to find the best one among $K$ actions. Summing over all states $S$ gives the bound $\Omega(\sqrt{SKT})$. However, this lower bound assumes the learner can observe the current state of the chosen arm. This information is unavailable in our setting, and the agent needs to detect a switch. As mentioned, up to logarithmic factors, $n_\epsilon = O(1/\epsilon^2)$ samples are needed to detect the change. Following the argument of~\cite{Auer19a} and assuming a sampling rate $p_s = 1$, the regret incurred before the change can be detected is $\Delta\sqrt{KT/L}/\epsilon$, for suboptimally $\Delta$ and minimum detectable change $\epsilon > \Delta$ for action $a$. Summing over all changes $L$ gives the contribution of regret of $\Delta\sqrt{KTL}/\epsilon$. Assuming that the information about the action reward distributions for each visited state is stored, the problem reduces to figuring out the current hidden state. The agent can use the collected history about the action rewards when the same state is revisited, not needing to relearn from scratch. Thus, the number of state switches $L$, becomes the number of switches into unique states $\Tilde{L}$, which can be much smaller than $L$. Adding both contributions and defining the minimum sub-optimally gap over all actions as $\Delta_{min}$ and minimum detectable change $\epsilon$ as $\epsilon_{min}$, gives the result.    
\end{proof}

We now consider the case where policies can exploit the context, therefore ``detecting'' changes in the reward distributions of actions. For the case where $p=0$, the context is a perfectly reliable way of detecting whether the state has changed. We use the result from ~\cite{Bouneffouf2020} for this scenario.

\begin{lemma}\label{theo:cb}

\citep{Bouneffouf2020} For any algorithm solving the CBCC problem
with context size $d$, with $(1 - p_{\nu})$ and $0 \leq p_{\nu} \leq 1$ there
exists a constant $\gamma> 0$, such that the lower bound of the
expected regret accumulated by the algorithm over $T$ iterations
is lower-bounded as follows: $E[R(T)] > \gamma \sqrt{Td}$. where $p_\nu$ is
the probability that the context is corrupted by an unknown
function $\nu$.
\end{lemma}

We arrive at our final result by combining the results from the analysis of context-free and contextual settings. For the CBCCAC setting, the lower bound depends on the context size $d$, the number of actions $K$ and states $S$, and the total number of state switches into unique states $\Tilde{L}$. Contextual or context-free cases then dominate the lower bound. We get for the lower bound.

\[
\EX[R(T)] \geq \min\Bigg\{\gamma\sqrt{Td},\sqrt{SKT} + \frac{\Delta_{min}\sqrt{KT\Tilde{L}}}{\epsilon_{min}} \Bigg\}.
\]

\subsection{Regret Upper Bound of COMBINE-UCB}

Our proposed algorithm combines two algorithms using a gradient bandit strategy. The upper bound of COMBINE depends on the algorithms used. We analyze COMBINE for the UCB instantiation, that is, COMBINE-UCB. It uses a gradient bandit to select between the LinUCB algorithm and a nonstationary UCB bandit algorithm we call NUCB for the remainder of this section. We carry out the analysis using the D-UCB algorithm for NUCB. D-UCB computes the discounted average action reward as $\bar{R}_{t}(a) = \sum_{\tau=0}^t \mathbbm{1}(J_\tau = a)\gamma^{t-\tau} r_\tau(a)$, where $J$ is the sequence of action choices up to time $t$ and $\mathbbm{1}$ is the indicator function. D-UCB also discounts the number of action plays computed as $N_{t}(a) = \sum_{\tau=0}^t \mathbbm{1}(J_\tau = a)\gamma^{t-\tau}$~\citep{Kocis2006}.  While our NUCB does not discount the number of action plays, it is possible to modify the exploration parameter $\alpha_B$ to a time-dependent version to achieve the same behavior as D-UCB. Let $\bar{R}_t'(a)$ and $\bar{R}_t^{''}(a)$ be the mean reward for action $a$ at time $t$ according to \ref{eq:avg_reward_sum} and D-UCB, respectively. Then, we can compute 

\begin{equation*}
    \alpha_B'(t) = \alpha_B \sqrt{\frac{\sum_{\tau=0}^t \mathbbm{1}(J_\tau = a)\gamma^{t-\tau}}{N_t(a)}} + \frac{\bar{R}'_t(a) - \bar{R}^{''}_t(a)}{\alpha_B \sqrt{\log(t)/\sum_{\tau=0}^t \mathbbm{1}(J_\tau = a)\gamma^{t-\tau}}}.
\end{equation*}
With this modified exploration parameter, we can rely on previous results. 

\begin{theorem}
The regret of COMBINE-UCB, using exploration paramter $\alpha_B'(t)$, is upper bounded with probability $1-\delta$ by
\[\EX[R(T)] \leq \sqrt{(\tau_1)d\log^3(KT\log(\tau_1)/\delta)} + \EX[\beta](T-\tau_1)^{(1+\eta)/2}\log(T-\tau_1),\]
with \[\EX[\beta] \leq K.\]
\end{theorem}

\begin{proof}
For the proof, we rely on the two following lemmas.

\begin{lemma}\label{lem:linucb} ~\citep{Chu2011} If SupLinUCB(Algorithm 3) is run with
\[
\alpha = \sqrt{\frac{1}{2} \log \frac{2TK}{\delta}},
\]
then with probability at least 1-$\delta$, the regret of the algorithm is 
\[
O\Bigg(\sqrt{Td\log^3(KT\log(T)/\delta)}\Bigg).
\]
\end{lemma}

\begin{lemma}\label{eq:lemmaducb}\citep{Garivier2011} Let $\gamma \in (1/2, 1)$. For any $T \geq 1$ and for any arm
$i \in \{1, . . . , K\}$ the number of sub-optimal arm choices using D-UCB is upper bounded by
\[
\EX[\Tilde{N}_T(i)] \leq C_1 T(1-\gamma)\log\frac{1}{1-\gamma} + C_2 \frac{\Upsilon_T}{1-\gamma} \log\frac{1}{1-\gamma},
\] for some constants $C_1$ and $C_2$
\end{lemma}

In the worst case, the upper regret bound of COMBINE-UCB is determined by the regret of the policy (LinUCB or NUCB) that performs worse in our problem setting, that is, $\EX[R(T)] \leq \max \{\EX[R_{LinUCB}(T)], \EX[R_{NUCB}(T)]\}$. Naturally, the gradient bandit will place more weight on the policy that exhibits better performance on average, so the regret typically falls between the regret of both policies. We can, therefore, express the upper bound by

\begin{equation}\label{eq:urb}
\EX[R(T)] \leq \EX[R_{LinUCB}(\tau_1)] + \EX[R_{NUCB}(\tau_2)],
\end{equation}

with $\tau_1$ being the number of time steps LinUCB is chosen, and $\tau_2:= T-\tau_1$ being the number of time steps the NUCB is chosen. Thus, in the asymptotic case, $\tau_1$  and $\tau_2 = T-\tau_1$ are determined by equation \ref{eq:equilib}. 

Knowing the horizon T and growth rate of the breakpoints $\Upsilon_T$, the discount factor can be chosen to minimize the RHS of lemma \ref{eq:lemmaducb}~\citep{Garivier2011}. Choosing the growth rate to be of order $\mathcal{O}(T^\eta)$ for some $\eta \in [0, 1)$, the regret is upper bounded by $T^{(1+\eta)/2} \log T$ for each action. For example, assuming $\eta$ is small, the number of breakpoints can be treated as constant over the horizon $T$ and can be interpreted as the number of state changes $L$.   

The NUCB policy explores using the adjacency matrix, restricting exploration to a subset of actions. The reach $\beta$ determines the size of the exploration action set, and its expected value depends on the average reward payoff of actions for each state and the policy. Therefore $\beta$ may improve regret performance by reducing the problem with $K$ actions into a problem with  $\EX[\beta] \leq K$ actions. Summing over all actions $\EX[\beta]$ gives the bound for NUCB. Inserting this result and the regret-bounds from Lemma \ref{lem:linucb} into \ref{eq:urb} gives the final result.
\end{proof}

\begin{remark} Results by~\citealp{Sutton2018} have demonstrated that the maximum convergence rate for a simple gradient bandit algorithm using the softmax policy is of order $1/t$. Therefore, the number of times the inferior strategy is selected converges with $\log(T)$ irrespective of the sub-optimality gap~\citep{Mei2020}. In contrast, our gradient bandit formulation converges to an optimal ratio, causing intermittent regret between LinUCB and NUCB.
\end{remark}

From this analysis, COMBINE using D-UCB matches the lower bound up to logarithmic factors when the gradient bandit settles on either $LinUCB$ or $NUCB$ eventually. Otherwise, if a dynamic equilibrium between the two policies is reached, we get an upper bound that matches the lower bound up to an additional $\sqrt{\EX[\beta]}$ factor, being $\sqrt{K}$ in the worst case.

\section{Experimentation}\label{sec:Exper}

 In the experimentation section, we describe the simulation environment and experimental setup for the simulation and real-world data sets.

\subsection{Simulation environment}\label{sec:Sim}
Our simulation environment is based on the medication adherence intervention scenario, where we need to recommend interventions for a set of users. We use this tool to generate artificial data under controlled settings, allowing us to gain an in-depth understanding of the performance of the proposed algorithm and compare it against state-of-the-art algorithms. The code is freely available on github\footnote{https://github.com/caisr-hh/CombineUserEnvironment} and follows the OpenAI gym API.

Our simulator maintains a continuous internal ``state'' $\omega$ for each user, which evolves according to a truncated random walk model defined on the domain $[0,1]$. We implemented the intuition of different ``levels'' or ``stages'' users might go through by discretizing the continuous space of $\omega$ into intervals, which define a finite space $\mathcal{S}$. 

More formally, given a set of actions $\mathcal{A}$ of cardinality $K$, we define a partition of the compact interval $Q=[0,1]$ on the real line such that there exists a sequence of sub-interval satisfying:

\begin{equation*}
    0 = q_0 < q_1 < \dots < q_{K-1} = 1,
\end{equation*}

where, for every subinterval $[q_k, q_{k+1})$, there exists a corresponding best action $a$. In other words, a mapping exists between action $a \in \mathcal{A}$ and a sub-interval of partition $q$, thus state space $\mathcal{S}$. Each subinterval has best actions $a^*_s = \argmax_{a\in \mathcal{A}} \mu_s(a, \cdot)$, which can be the same or different between states. In our simulator, the number of actions equals the number of states, but this is not a required assumption for the problem setting. We do not put any constraints on what action corresponds to what interval. However, for simplicity and interpretability, we assume that subinterval satisfies $[q_{k-1}, q_k]: a_k$, e.g., action $a_1$ corresponds to subinterval $[q_0, q_1)$, and so forth. Note that this mapping is unknown to the agent and needs to be learned by interacting with the environment.

The truncated random walk model simulates the evolution of $\omega$. It can be interpreted as a natural change in the ``health'' of the patient that may require specific interventions at particular ``levels''. This choice of state representation is based on computational convenience to generate contexts from $\omega$ directly. According to our problem setting, the context the agent receives at each iteration can be arbitrarily corrupted, making it useless for decision-making. In our simulation, we draw a random number from a uniform distribution on the interval $[0, 1]$ and assign it to $\omega$. 

More in line with the problem setting definition, it is perfectly possible to define the following: a transitions kernel containing the transition probabilities between states $\phi(s)$, context distribution $D_s$ and action reward $D_a(s)$ distribution for each action for a particular state $s$.

After the agent plays the action, the hidden state is updated. The function $\phi$ maps the current state $s_t$ to a state $s_{t+1}$ for time step $t+1$. $\phi$ in the simulator implements the process of updating $\omega$ and translating it into the next state. We do this in the following manner. At environment initialization, $T$ independent realizations are drawn from a Bernoulli distributed random variable $Z_t \sim {\mathcal {B}}(p)$ constituting set $\mathbf{Z}$, where $p$ denotes the probability of success, i.e., $P(Z = 1) = p$. For our simulations, we consider $p = 0.5$. Function $\phi(s_t)$ updates $s_t$. Let $f(\omega)$ be a function that maps $\omega$ to one of the subintervals, and $f^{-1}$ be its inverse. Then

\[
    \phi(s_t) = \left\{\begin{array}{lll}
         f(f^{-1}(s_t)+c) &\text{if } &Z_t = 1\\
         f(f^{-1}(s_t)-c) &&otherwise,
    \end{array}\right.
\]
    
for $\{Z_t \in \mathbf{Z}: t \in T\}$ and where $c \in [0,1]$ is a constant. $c$ is what we refer to as "state instability" in the experiments. $c$ determines the magnitude of change of $\omega$ in each time step and may be different for different users. 
Once chosen for a user, $c$ stays fixed for the whole simulation.

Distributing the action space over discrete intervals of the context space will lead to a nonlinear context-to-action-reward mapping. The action reward distribution changes abruptly when a different state is visited, resulting in a nonlinear ``step'' function from continuous state values to rewards. This mapping would result in the sub-par performance of the $CB$ algorithm since it assumes linear realizability of rewards. To allow linear models to perform well such that deciding between $CB$ and $MAB$ becomes nontrivial, we construct simple linear features of the context, i.e., we represent the context as the one-hot encoding of the best action given the corrupted context. Our simulation environment ``generates'' context vector $\mathbf{x}_{t, i}$ from $s_{t, i}$ by one-hot encoding the best action given the state $s_{t, i}$.
We use this simplification (unknown to the agent) only to increase the interpretability of results in the simulation environment. Neither the setting nor the algorithm depends on it, as shown in our real-world data experiments in Section~\ref{sec:Results}.

\subsection{Experimental Setup for the Simulation Environment}
\begin{figure}[tb]
  \begin{center}
    \includegraphics[width=.8\linewidth]{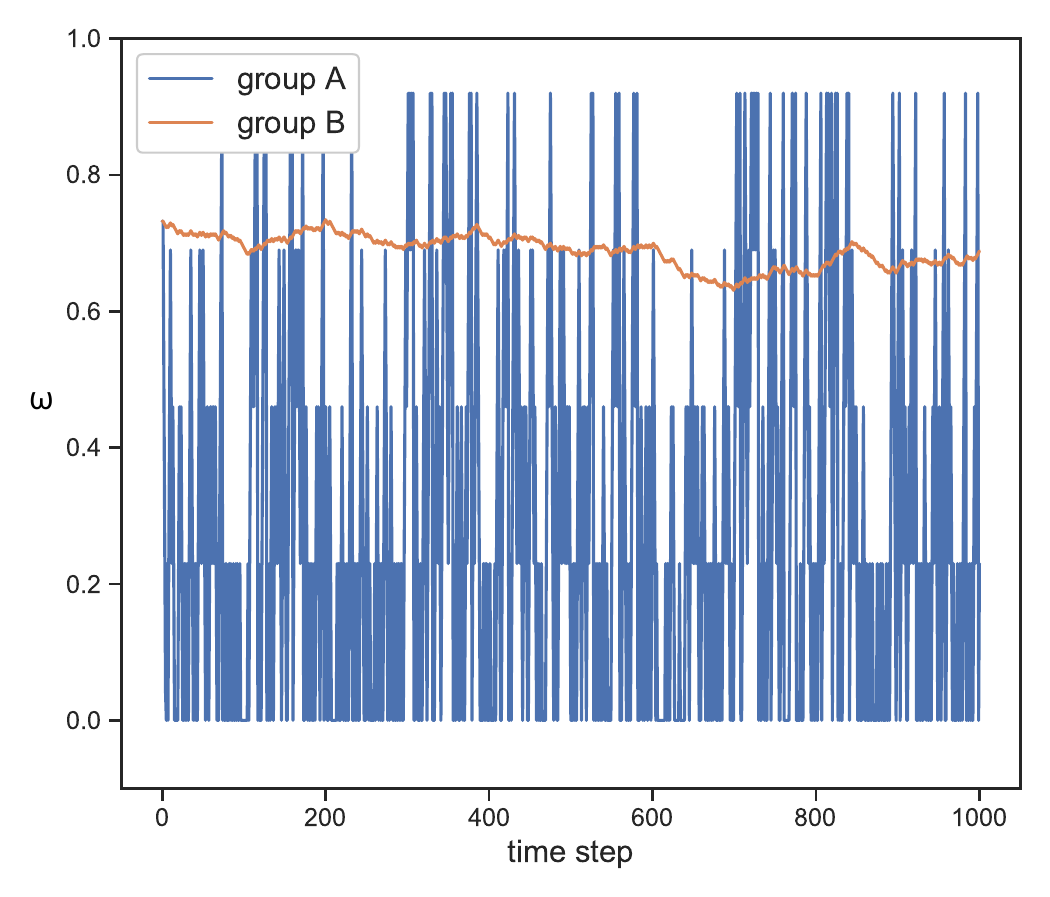}
  \end{center}
  \caption{Internal state ($\omega$) evolution for one user for each of the two groups. Users in group A exhibit a highly dynamic and rapidly changing state while having no corruption on the context (context corruption is not shown). The state of users in group B changes slowly, but their context is drawn from independent uniform noise.}
  \label{fig:state_evol_ex}
\end{figure}

First, we investigate the performance of the proposed algorithm on a set of users that exhibit extreme behaviors in terms of context corruption and fast state changes. Users in group A have no context corruption ($p_{\nu} = 0$) but change their hidden state rapidly ($c=0.1$), and group B provides a useless context ($p_{\nu} = 1$), but the hidden state (and best action) changes very slowly ($c=0.01$). An example of one user from each group is shown in figure~\ref{fig:state_evol_ex}. We expect the $CB$ to perform well in group A, while the $MAB_u$ would perform better in group B. The partition of $\omega$ is the same for all users in the simulation. Given this set of users, we investigate the performance of the algorithms as the number of actions increases. We run the algorithm on both groups at the same time. The regret is measured over 2500 time steps since the rank order of performances did not change for longer simulations. The challenge is to perform well on both groups of users.

\subsection{Experimental Setup for Real-world Data Sets}
\label{sec:datasets}

We investigate the performance of the proposed algorithm on real-world data sets from the UCI Machine Learning Repository\footnote{\url{https://archive.ics.uci.edu/ml/datasets.html}}. We primarily focused on time series or sequential data sets that exhibit ordering among classes (actions) in time.
We evaluate the proposed algorithm on these data sets, in particular: Activities of Daily Living Recognition Using Binary Sensors (ADL), Beijing Multi-Site Air-quality Data Set (BAQ), and Localization Data from Person Activity Data Set (Local).

For experimentation, we require the data to arrive in streams. We sample each context (features) sequentially. The agent chooses from the pool of actions (classes) at each round. We then reveal, through the reward, whether the instance was classified correctly ($r_t = 1$) or not ($r_t = 0$). We evaluate the algorithm's performance over various corruption levels using the same notion of regret defined in \ref{sec:ProbSett}.
All data sets underwent preprocessing. In the next paragraphs, we explain what we did for each data set in more detail.

\paragraph{Activities of Daily Living data set (ADLs)}
The ADLs data set contains activities and corresponding sensor events for two people (A and B) in an intelligent home environment. In particular, the recorded data contains timestamps of several user activities throughout the day. The data constitute approximately 320 and 500 samples for persons A and B. We re-sampled the data to every minute and padded the gaps between events by forward imputation to increase the number of samples. The final number of instances is 19517 for person A and 30208 for person B. The context is represented as a one-hot encoding of the sensor information. There are 9 Sensors for person B and 11 for person A. We combined (summed) the representation of sensor events that happened simultaneously during activity and capped the resulting vectors between 0 and 1. For context corruption, we draw a sensor event (index of a sensor) uniformly at random and generate the one-hot representation. We study the performance on the data set of each individual and combine the performance for the summary results. We run experiments with five different random seeds and ten random starting locations of the data in the first $3000$ time steps. The regret is evaluated for the first $15000$ time steps.

\paragraph{Beijing Multi-Site Air-quality Data Set (BAQ)}
This data set includes measurements from 12 nationally controlled air-quality monitoring sites from 2013-2017, recorded hourly, and all substations have the same number of instances. We run the algorithms on all 12 stations (corresponding to ``users'') simultaneously. We divided the particulate matter measurement (PM2.5) into five classes using equal frequency binning. We converted categorical features into their one-hot encoding. We used a random forest classifier to select the top $10$ predictive features, all numeric except the southwest (SW) wind direction, which is binary. For the corrupted context, similar to the ADL data set, we draw a random context from the uniform distribution with dimension $d=9$ on the domain $[0,1]$ and the SW wind direction from a binomial distribution with $p=0.5$. The data set constitutes approximately 35000 instances. We run experiments with five random seeds and evaluate the regret over all $35000$ time steps.

\paragraph{Localization Data from Person Activity Data Set (Local)}
The data set contains activities from five people performing a sequence of 11 activities every five times in a row. We combine the $x$, $y$, and $z$ coordinates of each of the four sensors into one set of features, constituting 12 features, and combine the data of all subjects (concatenation) for a total of approximately 35800 instances. As before, the corruption function constitutes a random context sampling from a uniform distribution on the domain $[0,1]$ with $d=12$. We run experiments with five different random seeds and ten random starting locations of the data in the first $3000$ time steps. The regret is measured over $30000$ time steps.

\subsection{Investigated Algorithms}

We compare the performance of the proposed algorithm with a stationary (\textbf{UCBBanditS}) and nonstationary (\textbf{UCBBanditNS}) UCB bandit, as well as two contextual bandits, \textbf{LinUCB}~\citep{Li2010} and \textbf{LinTS}~\citep{agrawal2014thompson}. UCBBanditS uses the UCB1 algorithm~\citep{Auer2002}. UCBBanditNS uses UCB1 but computes the average action rewards as a discounted sum according to equation \ref{eq:ducb}.

We also compare different combinations of $CB$ and $MAB$: LinUCB and Non-stationary UCB (\textbf{LinUCB + UCBBanditNS}), LinTS and Nonstationary UCB (\textbf{LinTS + UCBBanditNS}), both without reach parameter $\beta$. We test these against four versions of the meta-algorithm: \textbf{COMBINE-UCB}, \textbf{COMBINE-UCB common} (COMBINE-UCB with common adjacency matrix for all users), \textbf{COMBINE-softmax} (using a preference matrix for each user) and \textbf{COMBINE-softmax common} (a common preference matrix for all users). The algorithms that use different combinations of $CB$ and $MAB$ differ from COMBINE by not learning or using the adjacency matrix. They are otherwise identical, having a referee (gradient bandit) that chooses dynamically between $CB$ and $MAB$.

For all algorithms, we tuned the parameters to minimize the average regret over the respective time steps the experiments are run. For synthetic and real-world experiments, the parameters are: gradient bandit: $\delta_R = 0.5$,
LinUCB $\alpha = 1$, UCBBanditS/NS: $\alpha_B = 1$, $\gamma = 0.1$ and COMBINE-softmax: $\alpha_S = 10$. For LinTS, we set the algorithm-specific parameter $v=0.2$. All COMBINE versions use the same parameter values as the standalone algorithms. For all COMBINE versions, the adjacency matrix $\Lambda_i$ is updated according to the procedures in section \ref{sec:updateadjmatrix}.

\section{Results}\label{sec:Results}

In this section, we show the results of the experiments described above. We start with the simulated data, followed by the results on the real-world data sets.

\subsection{Simulation Analysis}
\begin{table}
 \caption{Regret performance on the simulated data. The regret is averaged over the number of actions for both group A and group B (mean $\pm$ std)}
  \centering
  \begin{tabular}{lccc}
    \toprule
    Method &Group A&Group B&Total\\
    \midrule
    UCBBanditNS &           1825 $\pm$ 584.3               &220.4 $\pm$ 148.2                       &1023 $\pm$ 913.0 \\
    UCBBanditS &         1714 $\pm$ 574.4               &850.3 $\pm$ 462.9                      &1282 $\pm$ 676.3 \\
    LinUCB &        33.31 $\pm$ 77.84              &1918 $\pm$ 357.73                       &975.8 $\pm$ 984.5 \\
    LinTS &         \textbf{14.73 $\pm$ 4.447 }             &1944 $\pm$ 324.46                       &979.8 $\pm$ 999.4 \\
    LinUCB+UCBBanditNS&     111.5 $\pm$ 103.4              &266.8 $\pm$ 184.2                      &189.1 $\pm$ 122.7 \\
    LinTS+UCBBanditNS&      69.32 $\pm$ 28.12              &273.0 $\pm$ 189.4                      &171.2 $\pm$ 169.0 \\
    \midrule
    COMBINE-UCB&         25.05 $\pm$ 6.452   &325.7 $\pm$ 237.9                      &175.4 $\pm$ 225.4 \\
    COMBINE-UCB common&   25.01 $\pm$ 6.253           &323.6 $\pm$ 237.7                      &174.3 $\pm$ 224.7 \\
    COMBINE-softmax&  24.89 $\pm$ 6.0            &213.0 $\pm$ 161.0                      &118.9 $\pm$ 147.5 \\
    COMBINE-softmax common& 25.01 $\pm$ 6.589      &\textbf{212.0 $\pm$ 155.5 }            &\textbf{118.7 $\pm$ 144.3 }\\
    \bottomrule
  \end{tabular}
  \label{tab:table1}
\end{table}

Table \ref{tab:table1} shows the regret over 2500 time steps of the investigated algorithms averaged over actions for both groups of users. We investigated 2, 5, 7, 9, 12, and 15 actions. For each combination of the parameters, we run five different random seeds and ten different initializations of the state per group, i.e., corresponding to ten users in each group. 

All algorithms utilizing the context performed well in group A, which is somewhat expected given that these users exhibit no context corruption. Therefore, context-to-action mapping is quickly learned. The regret on users B is significantly worse for the competing algorithms since the context is highly corrupted, prohibiting the learning of the expected reward for each action effectively. The combination approach works best on both groups of users as we dynamically adapt the chosen strategy based on how the MAB and CB perform during learning. The softmax variant performs the best overall since the next promising action is selected using the adjacency matrix compared to UCB, which might continue to explore and incur additional regret. 

\begin{figure}
  \centering
  \includegraphics[width=0.9\textwidth]{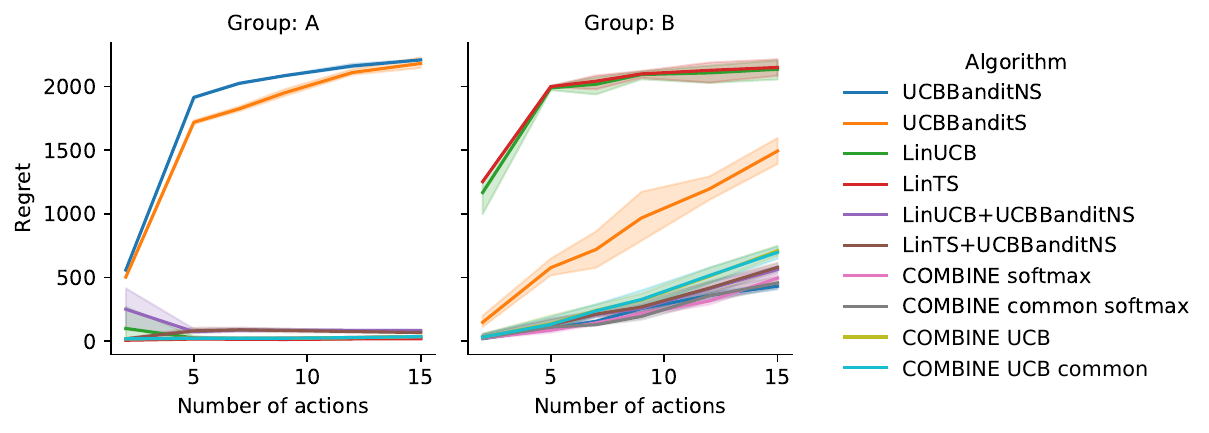}
  \includegraphics[width=0.9\textwidth]{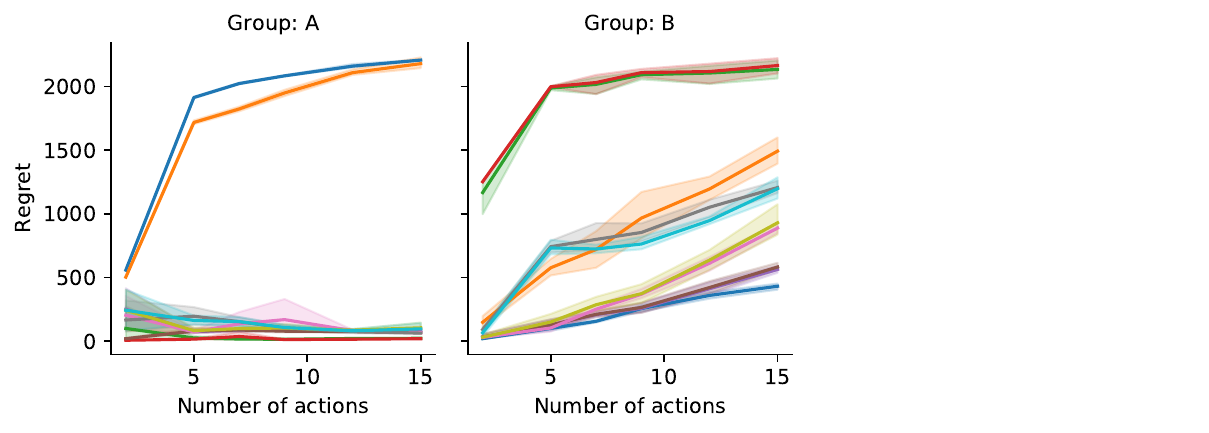}
  \caption{Regret for all investigated algorithms over actions and state for both groups A and B. Shaded areas show the standard deviation. (Top) An increase in the number of actions results in an expected increase in incurred regret for all algorithms not using the context and both groups A and B. (Bottom) The same experimental setup but COMBINE trains on all data without ignoring noisy data points. We observe that COMBINE is now more sensitive to the number of actions, particularly for group B.}
  \label{fig:regret_sim_log}
  \vspace{-2mm}
\end{figure}

Figure \ref{fig:regret_sim_log} shows the regret of the algorithms for our simulation study over a different number of actions.
In general, increasing the number of actions leads to an expected increase in regret in groups A and B. For group A, the single $CB$ and combination approaches are notable exceptions from this rule. An increase in the number of actions does not necessarily lead to strictly worse average regret. This is to some extent expected since the regret lower bound for contextual bandit problems scales with the dimensionality of the context space~\cite{Chu2011}, not with the number of actions. This property still holds when using the $ CB $ as base algorithms in combination approaches.

We note that combination approaches outperform single algorithms in Group B, exhibiting high context corruption but low state change. In group A, on the other hand, we observe that combining might only sometimes lead to the lowest incurred regret. We observe that LinTS outperforms COMBINE. We attribute this performance deficit to the referee, who must choose between the two base algorithms. This shows the potential limitations of COMBINE when in environments with low to little context noise, a concern we expect to play a less important role in practical applications where some form of context noise is ubiquitous. Furthermore, high state fluctuations effectively result in an i.i.d.-like sampling of the context, turning the problem into the standard contextual bandit setting. In such a setting, COMBINE may incur more regret due to the initial exploration of the referee, especially if there is little or no context noise.

\begin{figure}
    \centering
    \includegraphics[width=\textwidth]{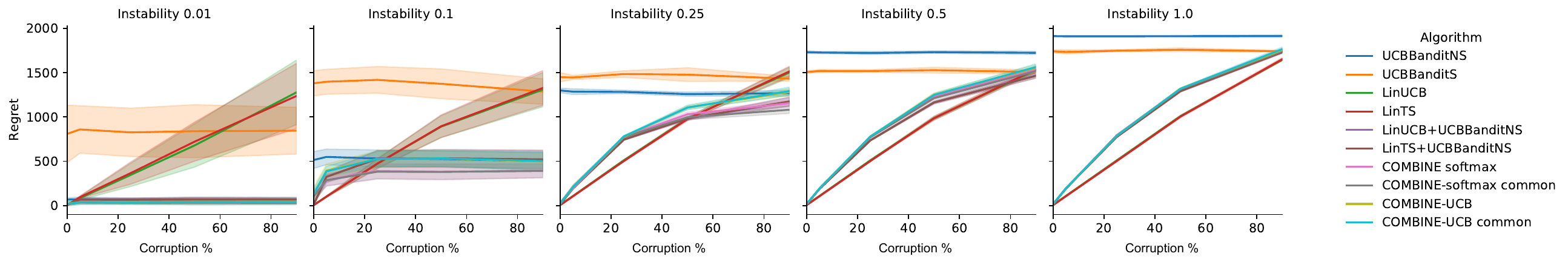}
    \caption{Regret over corruption levels and state instability for five actions. Notice the deteriorating performance of the nonstationary bandit as instability levels increase, providing a pseudo upper bound on the regret for the MAB components of the combination algorithms. A combination of high state instability (rapid change of actions within a time period) and high corruption levels leads to sub-par performance due to the dynamics of the referee. Shaded areas show standard deviation.}
    \label{fig:sim_stab_noise}
\end{figure}

To illustrate the referee's influence on regret, we ran additional experiments over a range of state instabilities (constant $c$ in section \ref{sec:Sim}) and corruption levels. In settings where a high level of corruption on contexts or state instability exists, choosing between $MAB_u$ or $CB$ might lead to higher overall regret, as both exhibit periods of better and worse performance; the referee biases action selection to one or the other intermittently. This "indecision" or switching by the referee comes with an additional cost where the pure $CB$ achieves lower regret in total compared to the combination. Figure \ref{fig:sim_stab_noise} illustrates this fact, showing the average regret for five actions. 

To investigate this point further, we explore the switching dynamics of the referee. We primarily focus on the simulation where all users exhibit the same levels of corruption and state instability, i.e., all users are the same. We observe that, in the setting of high instability and low corruption, the bandit is overtaken by the $CB$ as expected. However, even slight corruption or inaccuracies in learning the mapping between contexts and action rewards will lead to the $MAB_u$ exhibiting better performance. 

\begin{figure}
    \centering
    \includegraphics[width=\textwidth]{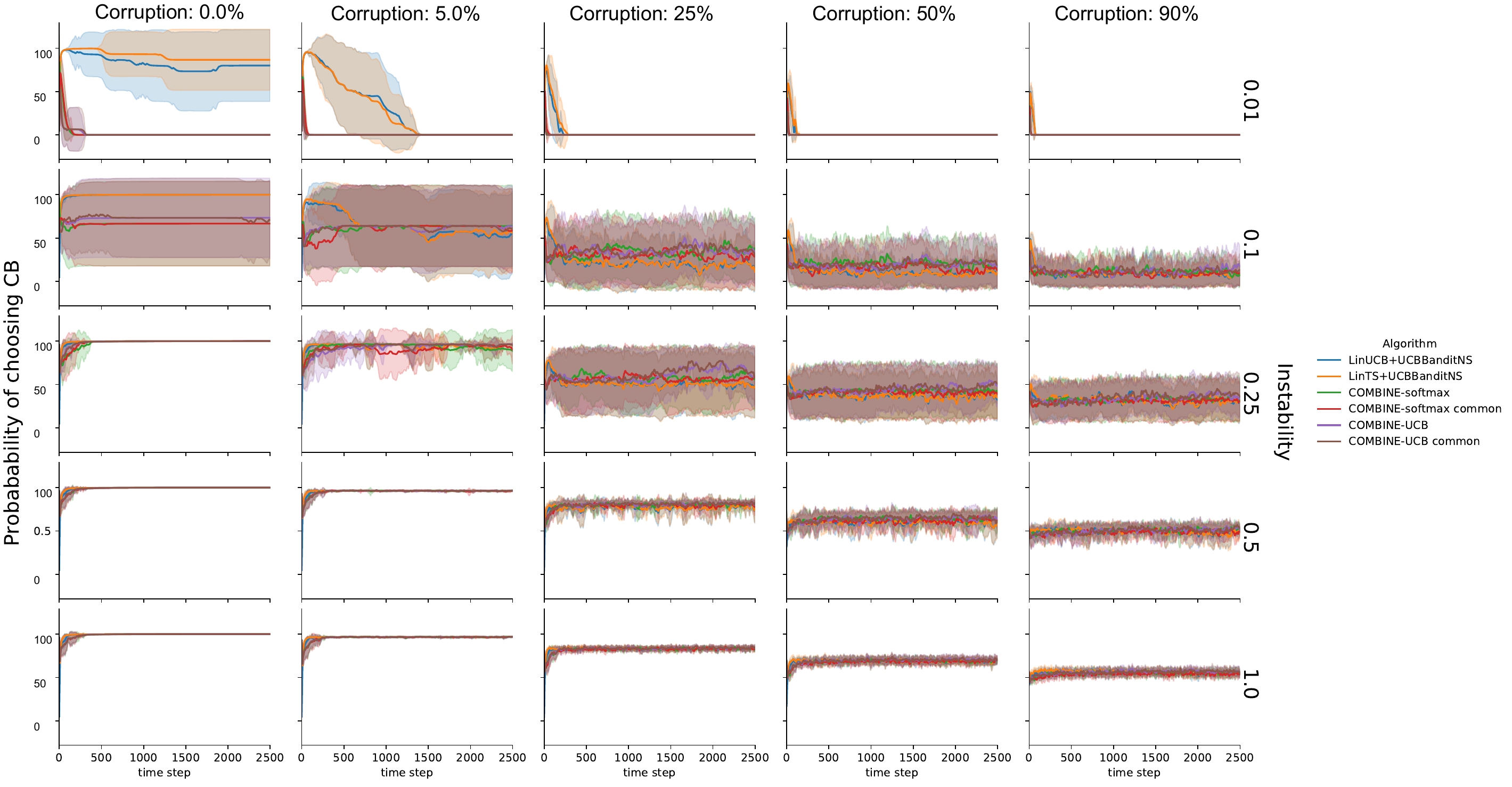}
    \caption{Probability of choosing $CB$ over time for different corruption levels and state instability. Results are shown for $10$ actions (thus states). The relative performance of both MAB and $CB$ determine the dynamic equilibrium between the two policies that are used for action selection.}
    \label{fig:ratio_sim}
    \vspace{-2mm}
\end{figure}

Figure \ref{fig:ratio_sim} shows the dynamics of the referee over time, corruption levels, and state stability for ten actions. The referee responds to changes in corruption levels or state instability quickly, adjusting its choice of the base algorithm according to the performance of either $CB$ or $MAB_u$ in the most recent time steps. We observe that the probability of choosing either the $CB$ or the $MAB_u$ falls into a dynamic equilibrium. This equilibrium exhibits higher variance in areas where either the $CB$ or $MAB_u$ show high short-term performance. In case of high state instability and corruption, neither $CB$ nor $MAB_u$ exhibits strictly higher performance on average. As a result, a combination of both does not yield superior performance compared to using a $CB$ or $MAB_u$ individually in the extreme case.  

Overall, at appropriate levels of corruption and state instability, the agents can decide between either contextual bandit or MAB, as one performs better than the other over time. For lower state instability, the referee settles for a dynamic equilibrium, which shows better regret performance than single approaches for a broad range of context corruption, particularly for higher levels of corruption.%In scenarios where corruption levels may change over time, hybrid methods can outperform non-combination approaches for the scenarios investigated.  

\subsection{Real-world Data Sets Analysis}

Table \ref{tab:table2} illustrates the average results for the three investigated data sets. As can be seen, our algorithm with the softmax bandit variant performs the best overall on the data sets when averaged over the corruption levels investigated. On the BAQ data set, we can also observe that the performance difference between the algorithms is the least significant compared to the ADL and Local data sets. This can be explained by high state permanence in both data sets (ADL and Local), allowing the agent to learn the state transition matrix effectively. The BAQ data set exhibits strong fluctuations between actions in time and nonlinear correlations between context and pollution level, violating the linear realizability assumption. This makes it quite tricky for the $CB$ and $MAB_u$ to perform well. We observe that both UCBBanditS and UCBBanditNS perform relatively poorly, and the $CB$s have a better chance of incurring less regret. We see that our algorithm with the softmax bandit variant performs the best for high corruption levels. We attribute the better performance to using the action transition probabilities instead of the upper confidence bound of action rewards, resulting in less exploration of actions that do not follow sequentially.

We applied a slight smoothing to the features and target variable (second-order Savitzky-Golay-filter, windows size 151). This improved the performance of all algorithms slightly without making the problem significantly easier for the $CB$ on average. The most significant change can be observed for the softmax bandit, where we see a reduction in regret by 50\% to 75\% depending on the corruption level compared to the non-smoothed version.  

\begin{table}
 \caption{Performance on real-world data sets. Average over corruption levels. The number in the parenthesis indicates the number of time steps the regret is measured over (mean $\pm$ std).}
  \centering
  \resizebox{\textwidth}{!}{\begin{tabular}{lcccc}
    \toprule
    Method  & ADL(15000)& Local(30000)& BAQ(35000)& BAQ smoothed(35000)\\
    \midrule
    UCBBandit                &9108 $\pm$ 481.0    &20690 $\pm$ 469      &26708 $\pm$ 412.0    &23419 $\pm$ 285.0    \\
 
    UCBBanditNS                  &4601 $\pm$ 1215    &21027 $\pm$ 59.00    &25202 $\pm$ 163.0    &19344 $\pm$ 37.00    \\
   
    LinUCB                 &4915 $\pm$ 2592    &13670 $\pm$ 3536   &19513 $\pm$ 278.0    &17013 $\pm$ 4358   \\

    LinTS                 &5058 $\pm$ 948.0      &13579 $\pm$ 3654  &20115 $\pm$ 281.0    &18774 $\pm$ 5020   \\

    LinUCB+UCBBanditNS              &4084 $\pm$ 2592  &10839 $\pm$ 1089  &19684 $\pm$ 133.0    &14267 $\pm$ 1862     \\
 
    LinTS+UCBBanditNS              &4121 $\pm$ 948.0   &10761 $\pm$ 1139  &19824 $\pm$ 254.0    &14309 $\pm$ 1901     \\
    
    \midrule
    COMBINE-UCB               &4189 $\pm$ 1064     &12806 $\pm$ 2763      &19709 $\pm$ 165.0    &15912 $\pm$ 2624     \\

    COMBINE-UCB common          &4171 $\pm$ 1015   &12762 $\pm$ 2638  &19657 $\pm$ 212.0    &15994 $\pm$ 2473   \\

    COMBINE-softmax        &1326 $\pm$112.0   &2784 $\pm$ 33.00  & 16106 $\pm$ 165.0    &3828 $\pm$ 33.0     \\

    COMBINE-softmax common  &\textbf{1295 $\pm$118.0 }  &\textbf{2655 $\pm$ 35.0 } &\textbf{15584 $\pm$ 154.0 }  &\textbf{3568 $\pm$ 60.0 }\\
 
    \bottomrule
  \end{tabular}}
  \label{tab:table2}
\end{table}

\begin{figure}
    \centering
    \captionsetup[subfigure]{justification=centering}
    \begin{subfigure}[b]{\textwidth}
        \includegraphics[width=\textwidth]{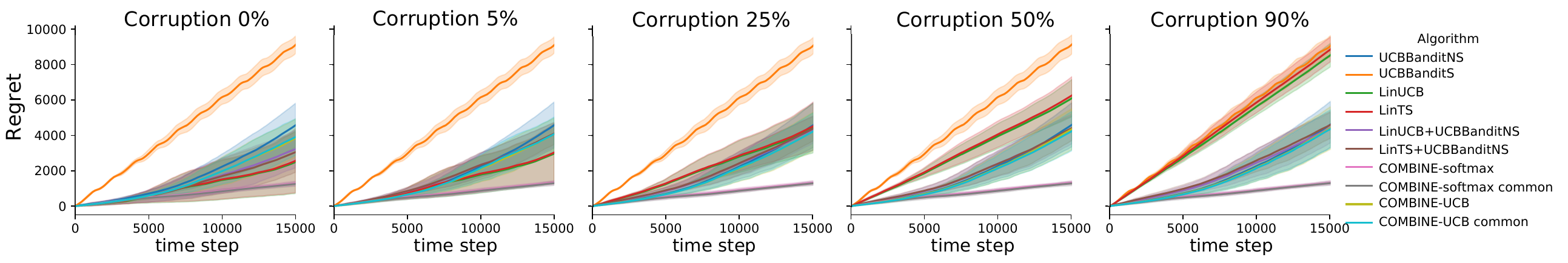}
        \caption{ADL data set.}
        \label{fig:regret_adlb_sub}
    \end{subfigure}
    \\
    \begin{subfigure}[b]{\textwidth}
        \includegraphics[width=\textwidth]{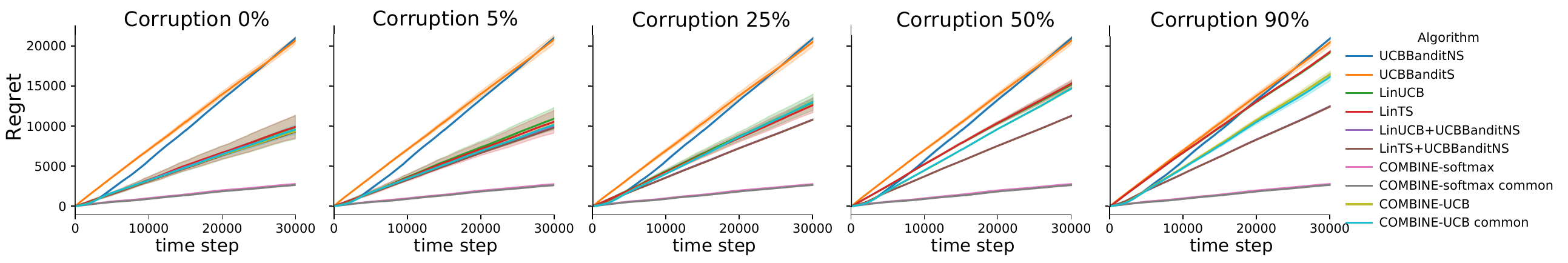}
        \caption{Local data set.}
        \label{fig:regret_local_sub}
    \end{subfigure}%
    \\
     \begin{subfigure}[b]{\textwidth}
        \includegraphics[width=\textwidth]{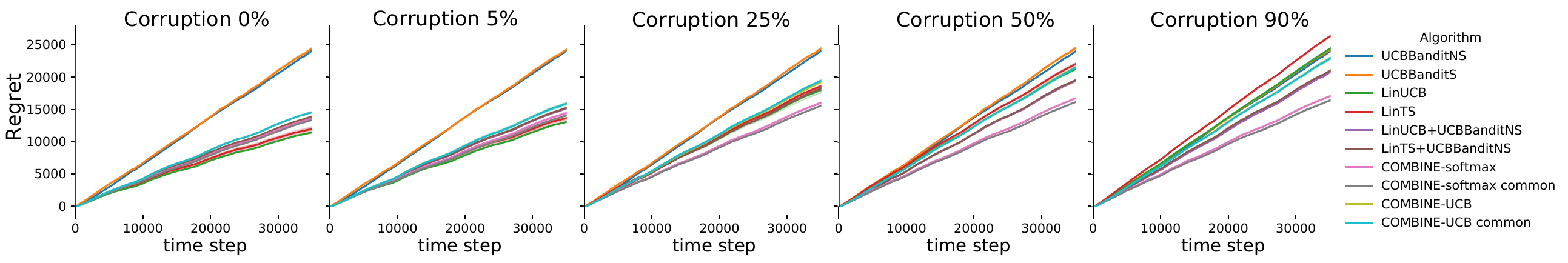}
        \caption{BQA data set.}
        \label{fig:regret_bqa_sub}
    \end{subfigure}%
    \\
    \begin{subfigure}[b]{\textwidth}
        \includegraphics[width=\textwidth]{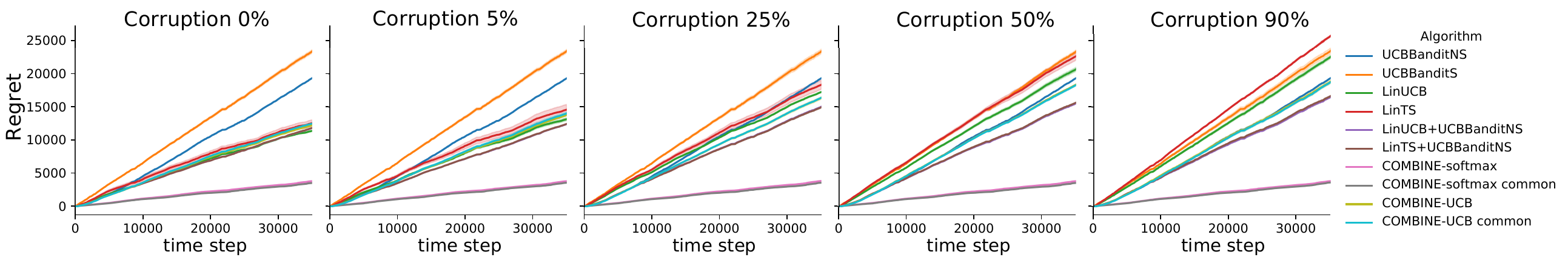}
        \caption{BQA smoothed data set.}
        \label{fig:regret_bqa_smooth_sub}
    \end{subfigure}%
    
    \caption{The softmax bandit version of COMBINE consistently outperforms other algorithms over all time steps. \ref{fig:regret_adlb_sub} Combined regret curves of both users in ADL data set. The softmax bandit version of our algorithm performs best mainly due to the good performance of the MAB component. \ref{fig:regret_local_sub} Similar to the ADL data set, the softmax bandit variant of our algorithm performs the best. The algorithm quickly convergences to using the soft-max bandit for action selection, incurring the least regret. \ref{fig:regret_bqa_sub} BAQ data set. For the non-smoothed version, the level of pollution can change rapidly between time steps, resulting in sub-par performance. Only for high corruption levels we observe the $MAB_u$ having an advantage. \ref{fig:regret_bqa_smooth_sub} Slight smoothing applied significantly improves performance. Shaded areas show standard deviation.}
    \label{fig:regret_realdata}
\end{figure}

Figure \ref{fig:regret_realdata} shows the regret curves of all algorithms over different levels of corruption of the context. For nearly all levels of corruption and all data sets, the softmax bandit variant of our algorithm outperforms other combination or non-combination algorithms consistently over time. Notable exceptions are on the BAQ data set, where we see a performance advantage only for higher state corruption levels above 5\% compared to LinTS or LinUCB. Neither $CB$ nor $MAB_u$ performs well, leading to a dynamic equilibrium of the referee similar to the simulations, resulting in additionally incurred regret for combination algorithms. The referee can easily decide between policies for higher corruption levels, and COMBINE-softmax can incur the least regret.

\begin{figure}
    \centering
    \includegraphics[width=\textwidth]{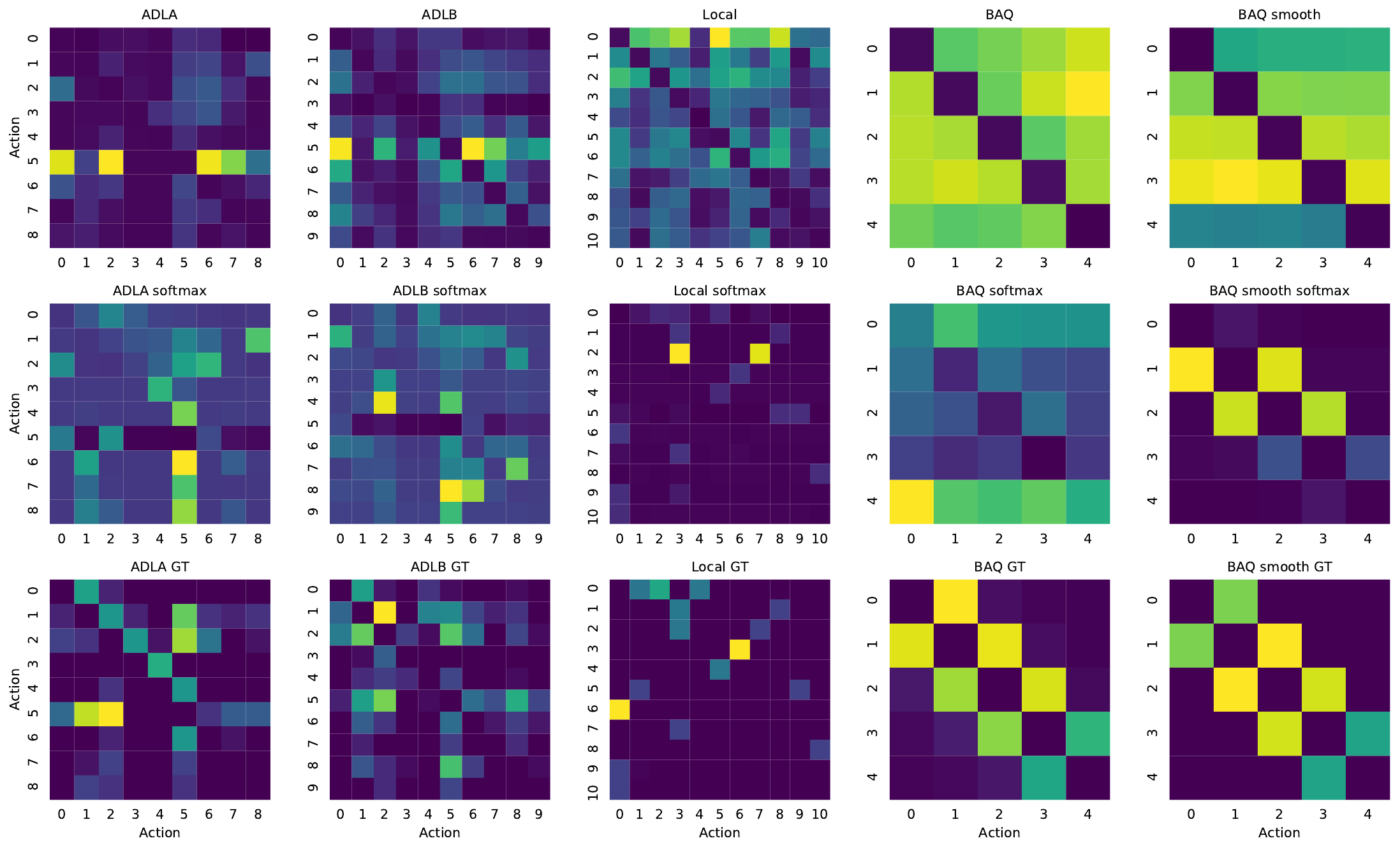}
    \caption{Adjacency Matrix for all data sets of COMBINE-UCB (top), COMBINE using the softmax bandit variation (middle) and the respective ground truth (GT) (bottom). Brighter areas indicate a higher number of transitions between two actions. The adjacency matrix has been split for the ADL data set into user A and user B. Specifically for the Local and BAQ smooth data sets, we observe that COMBINE-softmax learns the adjacency matrix more faithfully to the ground truth, leading to more targeted (fewer actions are included in $\mathcal{U}$) action selection and less regret incurred. A notable exception is the BAQ data set without smoothing, where COMBINE-softmax did not learn the adjacency matrix particularly well.}
    \label{fig:AdjecencyMatrix}
\end{figure}

Figure \ref{fig:AdjecencyMatrix} shows the learned adjacency matrix of the softmax bandit for all data sets. In most cases, using the softmax version leads to more sparse matrices and less exploration when choosing the subsequent promising actions. This works well for the ADL and Local data sets. The BQA data set exhibits significant fluctuations in the target variable and less pure action-to-action transition probabilities. Smoothing helps sparsify the adjacency matrix and, therefore, more predictable transitions, leading to significantly improved regret performance.        

\begin{figure}
    \centering
    \captionsetup[subfigure]{justification=centering}
    \begin{subfigure}[b]{\textwidth}
        \includegraphics[width=\textwidth]{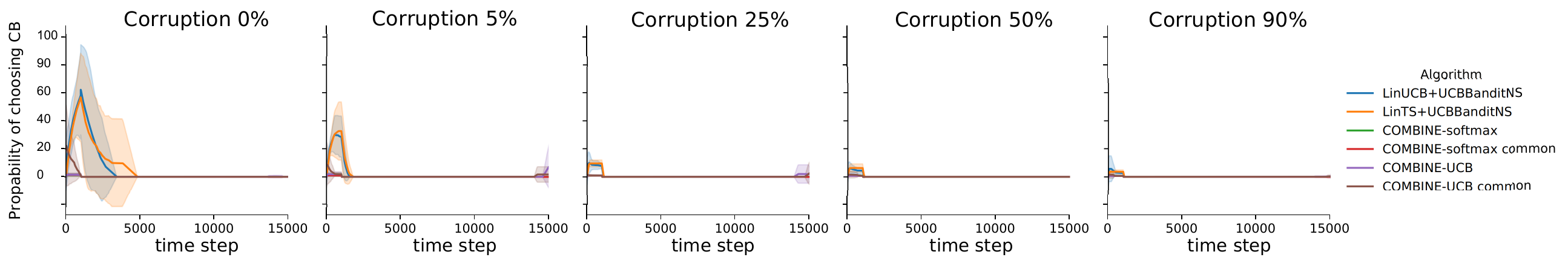}
        \caption{ADL data set. Both users combined.}
        \label{fig:adl_ratio_sub}
     \end{subfigure}%
    \\
     
    \begin{subfigure}[b]{\textwidth}
        \includegraphics[width=\textwidth]{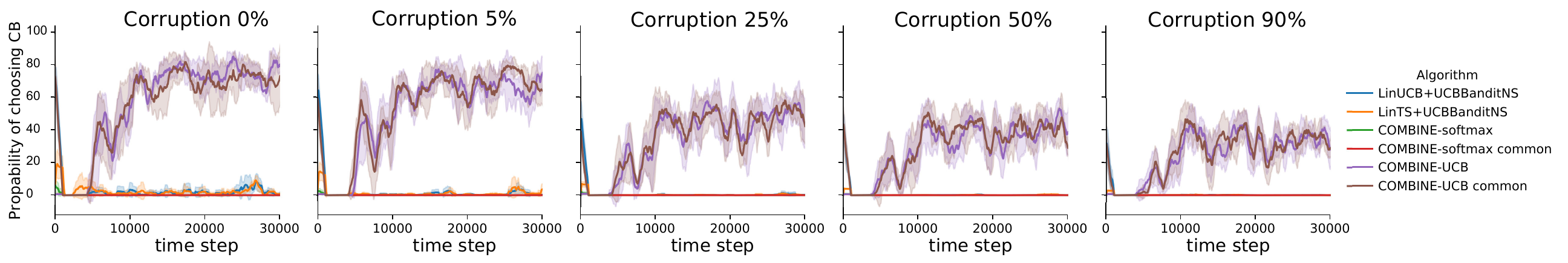}
        \caption{Local data set.}
        \label{fig:local_ratio_sub}
     \end{subfigure}%
    \\
     \begin{subfigure}[b]{\textwidth}
        \includegraphics[width=\textwidth]{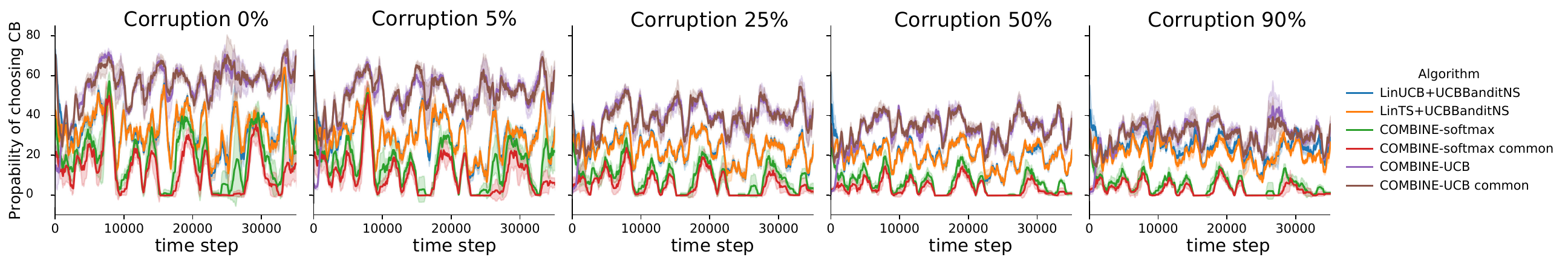}
        \caption{BAQ data. All stations combined.}
        \label{fig:bqa_ratio_sub}
     \end{subfigure}%
    \\
    \begin{subfigure}[b]{\textwidth}
        \includegraphics[width=\textwidth]{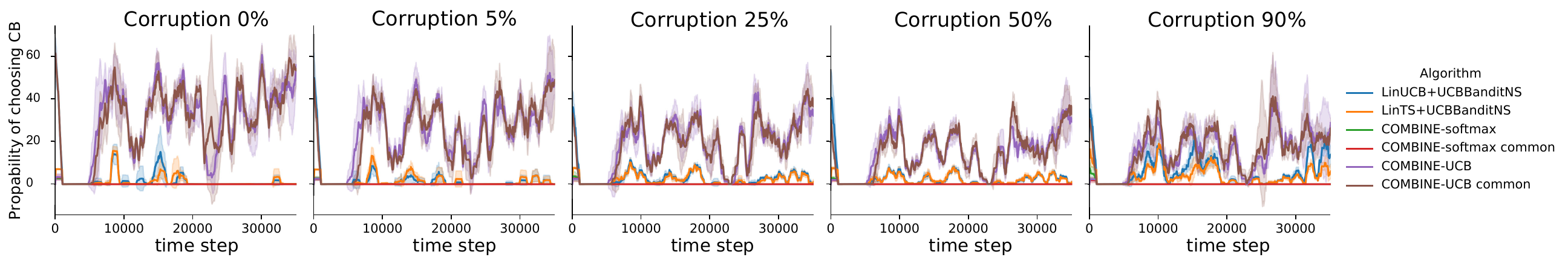}
        \caption{BAQ data set smoothed version. All stations combined.}
        \label{fig:bqa_smooth_ratio_sub}
    \end{subfigure}%
    
    \caption{Probability of choosing the $CB$ over the MAB for the different real-world data sets. \ref{fig:adl_ratio_sub} The ADL data almost universally favor the $MA$, with the stationary $MAB$ experiencing periods where the $CB$ is preferred, mostly in regions of action transition. \ref{fig:local_ratio_sub} The $MAB$ approach outperforms the $CB$ quickly on the Local data set. As a failure case, the MAB component of COMBINE-UCB does not perform well, resulting in additional regret. \ref{fig:bqa_ratio_sub} Neither $CB$ nor $MAB$ performs exceptionally well, resulting in fluctuating behavior of the referee. \ref{fig:bqa_smooth_ratio_sub} Smoothing the BAQ data set reduces sporadic action changes and allows the $MAB$ to outperform the $CB$. Shaded areas show standard deviation.}
    \label{fig:ratio_real}
\end{figure}

Figure \ref{fig:ratio_real} shows the dynamics of the referee on the real-world data sets. The ADL data set exhibits strong ordering and low action change frequency such that the $MAB$ variants of the algorithms can achieve low regret without knowing the context. This is further exacerbated when adding corruption to the context, diminishing the performance of the $CB$. 

Similarly to the ADL dataset, the $MAB$ outperforms the $CB$ early on in the Local data set. Our algorithm's non-softmax version performs comparatively worse on this data set due to the $MAB_u$ component being non-competitive. We attribute this to the fact that the $MAB_u$ component still uses UCB to select actions. Given that we do not update the $MAB_u$ if it was not chosen by the referee for actions selection, UCB might explore more often, incurring additional regret and increasing the reach parameter $\beta$ rapidly. This may effectively lead to (i) the whole actions space being available for exploration and (ii) the "over-exploitation" of actions that, due to $\beta$, rarely have been tried before. These actions receive a large exploration bonus due to the high upper-confidence-bound estimate (line 24 in algorithm \ref{alg:alg2.1}), and it may require a significant number of trials to reduce said bonus. 

We observe a significant change in the referee's behavior for all algorithms on the BAQ data set. Contexts (thus actions) can frequently change. Coupled with the nonlinear context that does not allow a good mapping from contexts to actions, the performance of all algorithms suffers. Interestingly, the softmax version provides enough advantage over the $CB$ to be chosen more frequently, further amplified by corrupting the context. Smoothing the data set allows the $MAB_u$ to overtake the $CB$ rapidly, reducing regret, particularly for the softmax variation of our algorithm.

\subsubsection{Tuning \texorpdfstring{$\alpha_B$}{} and \texorpdfstring{$\alpha_S$}{}}

As we have mentioned in earlier sections, the non-softmax $MAB_u$ components explore actions in a manner that does not allow competitive regret performance to the softmax variant. All combination algorithms have a UCB $MAB$ in their core component, so we need to tune the $\alpha_B$ parameter for optimal exploration and exploitation. We briefly investigate the effect of the exploration parameter $\alpha_B$ to see if we can improve upon regret.

\begin{figure}
    \centering
    \includegraphics[width=\textwidth]{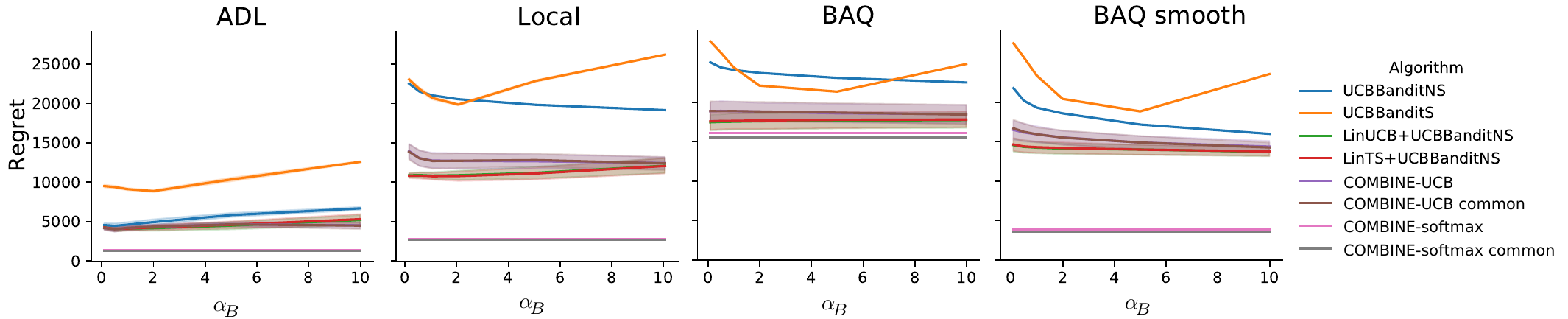}
    \caption{Regret over a range of $\alpha_B$ values of the $MAB$ component of each algorithm. Each data set has a unique optimal value. The softmax bandit version performs best even after tuning the other $MAB$ agents. Shaded areas show standard deviation.}
    \label{fig:ucb_test}
\end{figure}

Figure \ref{fig:ucb_test} shows the regret achieved over a range of values of the parameter for the investigated data sets. The performance of the algorithms slightly improves for an optimal $\alpha_B$ unique to each data set. Even with tuning, the softmax bandit variation of our algorithm outperforms all other investigated methods with the added benefit of not needing any tuning. Even when tuning the exploration rate from very high to very low exploration, we do not significantly improve performance such that we are on par with the softmax $MAB_u$ variant.

 \begin{figure}
    \centering
    \includegraphics[width=\textwidth]{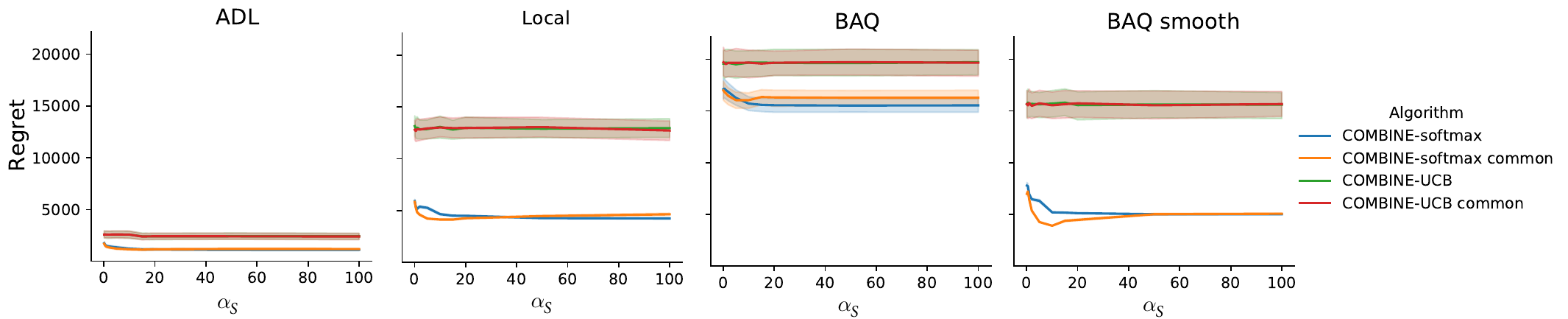}
    \caption{Regret over a range of $\alpha_S$ values for algorithms using the adjacency matrix. Best results are obtained for sufficiently large $\alpha_S$, e.g., in $5-10$ range.}
    \label{fig:alphas_test}
\end{figure}

Figure \ref{fig:alphas_test} shows the regret for various values of $\alpha_S$ for the algorithms that use the adjacency matrix for action selection. The most significant effect on regret can be observed for low values between $0-20$. The softmax-bandit variant of COMBINE shows the highest sensitivity to this parameter, which is not surprising since updates to the preference matrix can significantly affect the resulting probability distribution over actions. Biasing action selection towards promising next actions with fewer iterations effectively results in more ``greedy'' action selection and less exploration. The effect on the COMBINE-UCB version of the algorithm is less significant given that the ranking of the actions is not influenced as strongly by $\alpha_S$.

\section{Conclusions}\label{sec:Conclu}

We introduced a sequential decision-making problem setting where the observed corrupted contexts are determined by an underlying state that obeys the Markov property. This state evolution allows the incorporation of knowledge about the order of actions used to guide the exploration and exploitation of actions. In a novel way, we combine two previously investigated settings, namely the Contextual Bandit with Corrupted Contexts and Regime Switching Bandits. In our setting, the agent needs to balance the exploration and exploitation of actions based on its belief about the reliability of information from learned state transitions and context information. We propose a meta-algorithm called COMBINE and an instantiation based on an upper-confidence bound combined with one-step transition probabilities. 

We show improved performance compared to contemporary methods through empirical evaluation of the algorithm on simulated and real-world data, mainly when using the softmax variant of COMBINE. On simulated data, for extreme cases of high state fluctuations and low context corruption, or, alternatively, low state fluctuations and high context corruption, using single approaches, i.e., either a $ CB $ or $ MAB $, might minimize incurred regret. In more realistic settings, however, when there is enough exploitable information for the agent to make decisions, the COMBINE approach significantly outperforms single methods.  

\section{Ackowledgement}

Funding: The authors thank Vinnova for financing the research work under
the project iMedA, under grant number: 2017-04617.

\bibliographystyle{apacite}
\bibliography{references}  %%% Remove comment to use the external .bib file (using bibtex).

\begin{thebibliography}{}

\bibitem [\protect \citeauthoryear {%
Abouserie%
}{%
Abouserie%
}{%
{\protect \APACyear {1994}}%
}]{%
Reda1994}
\APACinsertmetastar {%
Reda1994}%
\begin{APACrefauthors}%
Abouserie, R.%
\end{APACrefauthors}%
\unskip\
\newblock
\APACrefYearMonthDay{1994}{}{}.
\newblock
{\BBOQ}\APACrefatitle {Sources and Levels of Stress in Relation to Locus of
  Control and Self Esteem in University Students} {Sources and levels of stress
  in relation to locus of control and self esteem in university
  students}.{\BBCQ}
\newblock
\APACjournalVolNumPages{Educational Psychology}{14}{3}{323-330}.
\newblock
\begin{APACrefURL} \url{https://doi.org/10.1080/0144341940140306}
  \end{APACrefURL}
\newblock
\begin{APACrefDOI} \doi{10.1080/0144341940140306} \end{APACrefDOI}
\PrintBackRefs{\CurrentBib}

\bibitem [\protect \citeauthoryear {%
Agarwal%
, Luo%
, Neyshabur%
\BCBL {}\ \BBA {} Schapire%
}{%
Agarwal%
\ \protect \BOthers {.}}{%
{\protect \APACyear {2016}}%
}]{%
Agarwal16}
\APACinsertmetastar {%
Agarwal16}%
\begin{APACrefauthors}%
Agarwal, A.%
, Luo, H.%
, Neyshabur, B.%
\BCBL {}\ \BBA {} Schapire, R\BPBI E.%
\end{APACrefauthors}%
\unskip\
\newblock
\APACrefYearMonthDay{2016}{}{}.
\newblock
{\BBOQ}\APACrefatitle {Corralling a Band of Bandit Algorithms} {Corralling a
  band of bandit algorithms}.{\BBCQ}
\newblock
\APACjournalVolNumPages{CoRR}{abs/1612.06246}{}{}.
\newblock
\begin{APACrefURL} \url{http://arxiv.org/abs/1612.06246} \end{APACrefURL}
\PrintBackRefs{\CurrentBib}

\bibitem [\protect \citeauthoryear {%
Agrawal%
\ \BBA {} Goyal%
}{%
Agrawal%
\ \BBA {} Goyal%
}{%
{\protect \APACyear {2013}}%
{\protect \APACexlab {{\protect \BCnt {1}}}}}]{%
Agrawal2013}
\APACinsertmetastar {%
Agrawal2013}%
\begin{APACrefauthors}%
Agrawal, S.%
\BCBT {}\ \BBA {} Goyal, N.%
\end{APACrefauthors}%
\unskip\
\newblock
\APACrefYearMonthDay{2013{\protect \BCnt {1}}}{29 Apr--01 May}{}.
\newblock
{\BBOQ}\APACrefatitle {Further Optimal Regret Bounds for Thompson Sampling}
  {Further optimal regret bounds for thompson sampling}.{\BBCQ}
\newblock
\BIn{} C\BPBI M.~Carvalho\ \BBA {} P.~Ravikumar\ (\BEDS), \APACrefbtitle
  {Proceedings of the Sixteenth International Conference on Artificial
  Intelligence and Statistics} {Proceedings of the sixteenth international
  conference on artificial intelligence and statistics}\ (\BVOL~31, \BPGS\
  99--107).
\newblock
\APACaddressPublisher{Scottsdale, Arizona, USA}{PMLR}.
\newblock
\begin{APACrefURL} \url{https://proceedings.mlr.press/v31/agrawal13a.html}
  \end{APACrefURL}
\PrintBackRefs{\CurrentBib}

\bibitem [\protect \citeauthoryear {%
Agrawal%
\ \BBA {} Goyal%
}{%
Agrawal%
\ \BBA {} Goyal%
}{%
{\protect \APACyear {2013}}%
{\protect \APACexlab {{\protect \BCnt {2}}}}}]{%
agrawal2014thompson}
\APACinsertmetastar {%
agrawal2014thompson}%
\begin{APACrefauthors}%
Agrawal, S.%
\BCBT {}\ \BBA {} Goyal, N.%
\end{APACrefauthors}%
\unskip\
\newblock
\APACrefYearMonthDay{2013{\protect \BCnt {2}}}{17--19 Jun}{}.
\newblock
{\BBOQ}\APACrefatitle {Thompson Sampling for Contextual Bandits with Linear
  Payoffs} {Thompson sampling for contextual bandits with linear
  payoffs}.{\BBCQ}
\newblock
\BIn{} S.~Dasgupta\ \BBA {} D.~McAllester\ (\BEDS), (\BVOL~28, \BPGS\
  127--135).
\newblock
\APACaddressPublisher{Atlanta, Georgia, USA}{PMLR}.
\newblock
\begin{APACrefURL} \url{http://proceedings.mlr.press/v28/agrawal13.html}
  \end{APACrefURL}
\PrintBackRefs{\CurrentBib}

\bibitem [\protect \citeauthoryear {%
Auer%
}{%
Auer%
}{%
{\protect \APACyear {2003}}%
}]{%
Auer2003b}
\APACinsertmetastar {%
Auer2003b}%
\begin{APACrefauthors}%
Auer, P.%
\end{APACrefauthors}%
\unskip\
\newblock
\APACrefYearMonthDay{2003}{{\APACmonth{03}}}{}.
\newblock
{\BBOQ}\APACrefatitle {Using Confidence Bounds for Exploitation-Exploration
  Trade-Offs} {Using confidence bounds for exploitation-exploration
  trade-offs}.{\BBCQ}
\newblock
\APACjournalVolNumPages{J. Mach. Learn. Res.}{3}{null}{397–422}.
\PrintBackRefs{\CurrentBib}

\bibitem [\protect \citeauthoryear {%
Auer%
, Cesa-Bianchi%
\BCBL {}\ \BBA {} Fischer%
}{%
Auer%
\ \protect \BOthers {.}}{%
{\protect \APACyear {2002}}%
}]{%
Auer2002}
\APACinsertmetastar {%
Auer2002}%
\begin{APACrefauthors}%
Auer, P.%
, Cesa-Bianchi, N.%
\BCBL {}\ \BBA {} Fischer, P.%
\end{APACrefauthors}%
\unskip\
\newblock
\APACrefYearMonthDay{2002}{05}{}.
\newblock
{\BBOQ}\APACrefatitle {Finite-time Analysis of the Multiarmed Bandit Problem}
  {Finite-time analysis of the multiarmed bandit problem}.{\BBCQ}
\newblock
\APACjournalVolNumPages{Machine Learning}{47}{}{235-256}.
\newblock
\begin{APACrefDOI} \doi{10.1023/A:1013689704352} \end{APACrefDOI}
\PrintBackRefs{\CurrentBib}

\bibitem [\protect \citeauthoryear {%
Auer%
, Cesa-Bianchi%
, Freund%
\BCBL {}\ \BBA {} Schapire%
}{%
Auer%
\ \protect \BOthers {.}}{%
{\protect \APACyear {2003}}%
}]{%
Auer2003}
\APACinsertmetastar {%
Auer2003}%
\begin{APACrefauthors}%
Auer, P.%
, Cesa-Bianchi, N.%
, Freund, Y.%
\BCBL {}\ \BBA {} Schapire, R\BPBI E.%
\end{APACrefauthors}%
\unskip\
\newblock
\APACrefYearMonthDay{2003}{{\APACmonth{01}}}{}.
\newblock
{\BBOQ}\APACrefatitle {The Nonstochastic Multiarmed Bandit Problem} {The
  nonstochastic multiarmed bandit problem}.{\BBCQ}
\newblock
\APACjournalVolNumPages{SIAM J. Comput.}{32}{1}{48–77}.
\newblock
\begin{APACrefURL} \url{https://doi.org/10.1137/S0097539701398375}
  \end{APACrefURL}
\newblock
\begin{APACrefDOI} \doi{10.1137/S0097539701398375} \end{APACrefDOI}
\PrintBackRefs{\CurrentBib}

\bibitem [\protect \citeauthoryear {%
Auer%
, Gajane%
\BCBL {}\ \BBA {} Ortner%
}{%
Auer%
\ \protect \BOthers {.}}{%
{\protect \APACyear {2019}}%
}]{%
Auer19a}
\APACinsertmetastar {%
Auer19a}%
\begin{APACrefauthors}%
Auer, P.%
, Gajane, P.%
\BCBL {}\ \BBA {} Ortner, R.%
\end{APACrefauthors}%
\unskip\
\newblock
\APACrefYearMonthDay{2019}{25--28 Jun}{}.
\newblock
{\BBOQ}\APACrefatitle {Adaptively Tracking the Best Bandit Arm with an Unknown
  Number of Distribution Changes} {Adaptively tracking the best bandit arm with
  an unknown number of distribution changes}.{\BBCQ}
\newblock
\BIn{} A.~Beygelzimer\ \BBA {} D.~Hsu\ (\BEDS), \APACrefbtitle {Proceedings of
  the Thirty-Second Conference on Learning Theory} {Proceedings of the
  thirty-second conference on learning theory}\ (\BVOL~99, \BPGS\ 138--158).
\newblock
\APACaddressPublisher{Phoenix, USA}{PMLR}.
\newblock
\begin{APACrefURL} \url{http://proceedings.mlr.press/v99/auer19a.html}
  \end{APACrefURL}
\PrintBackRefs{\CurrentBib}

\bibitem [\protect \citeauthoryear {%
Baltrunas%
, Church%
, Karatzoglou%
\BCBL {}\ \BBA {} Oliver%
}{%
Baltrunas%
\ \protect \BOthers {.}}{%
{\protect \APACyear {2015}}%
}]{%
Baltrunas2015}
\APACinsertmetastar {%
Baltrunas2015}%
\begin{APACrefauthors}%
Baltrunas, L.%
, Church, K.%
, Karatzoglou, A.%
\BCBL {}\ \BBA {} Oliver, N.%
\end{APACrefauthors}%
\unskip\
\newblock
\APACrefYearMonthDay{2015}{}{}.
\newblock
{\BBOQ}\APACrefatitle {Frappe: Understanding the Usage and Perception of Mobile
  App Recommendations In-The-Wild} {Frappe: Understanding the usage and
  perception of mobile app recommendations in-the-wild}.{\BBCQ}
\newblock
\APACjournalVolNumPages{CoRR}{abs/1505.03014}{}{}.
\newblock
\begin{APACrefURL} \url{http://arxiv.org/abs/1505.03014} \end{APACrefURL}
\PrintBackRefs{\CurrentBib}

\bibitem [\protect \citeauthoryear {%
Bastani%
\ \BBA {} Bayati%
}{%
Bastani%
\ \BBA {} Bayati%
}{%
{\protect \APACyear {2020}}%
}]{%
Bastani2020}
\APACinsertmetastar {%
Bastani2020}%
\begin{APACrefauthors}%
Bastani, H.%
\BCBT {}\ \BBA {} Bayati, M.%
\end{APACrefauthors}%
\unskip\
\newblock
\APACrefYearMonthDay{2020}{}{}.
\newblock
{\BBOQ}\APACrefatitle {Online Decision Making with High-Dimensional Covariates}
  {Online decision making with high-dimensional covariates}.{\BBCQ}
\newblock
\APACjournalVolNumPages{Operations Research}{68}{1}{276-294}.
\newblock
\begin{APACrefURL} \url{https://doi.org/10.1287/opre.2019.1902}
  \end{APACrefURL}
\newblock
\begin{APACrefDOI} \doi{10.1287/opre.2019.1902} \end{APACrefDOI}
\PrintBackRefs{\CurrentBib}

\bibitem [\protect \citeauthoryear {%
Boldrini%
\ \protect \BOthers {.}}{%
Boldrini%
\ \protect \BOthers {.}}{%
{\protect \APACyear {2018}}%
}]{%
Boldrini2018}
\APACinsertmetastar {%
Boldrini2018}%
\begin{APACrefauthors}%
Boldrini, S.%
, De~Nardis, L.%
, Caso, G.%
, Le, M.%
, Fiorina, J.%
\BCBL {}\ \BBA {} Di~Benedetto, M\BHBI G.%
\end{APACrefauthors}%
\unskip\
\newblock
\APACrefYearMonthDay{2018}{Jan}{}.
\newblock
{\BBOQ}\APACrefatitle {muMAB: A Multi-Armed Bandit Model for Wireless Network
  Selection} {mumab: A multi-armed bandit model for wireless network
  selection}.{\BBCQ}
\newblock
\APACjournalVolNumPages{Algorithms}{11}{2}{13}.
\newblock
\begin{APACrefURL} \url{http://dx.doi.org/10.3390/a11020013} \end{APACrefURL}
\newblock
\begin{APACrefDOI} \doi{10.3390/a11020013} \end{APACrefDOI}
\PrintBackRefs{\CurrentBib}

\bibitem [\protect \citeauthoryear {%
Bouneffouf%
}{%
Bouneffouf%
}{%
{\protect \APACyear {2021}}%
}]{%
Bouneffouf2020}
\APACinsertmetastar {%
Bouneffouf2020}%
\begin{APACrefauthors}%
Bouneffouf, D.%
\end{APACrefauthors}%
\unskip\
\newblock
\APACrefYearMonthDay{2021}{}{}.
\newblock
{\BBOQ}\APACrefatitle {Corrupted Contextual Bandits: Online Learning with
  Corrupted Context} {Corrupted contextual bandits: Online learning with
  corrupted context}.{\BBCQ}
\newblock
\BIn{} \APACrefbtitle {ICASSP 2021 - 2021 IEEE International Conference on
  Acoustics, Speech and Signal Processing (ICASSP)} {Icassp 2021 - 2021 ieee
  international conference on acoustics, speech and signal processing
  (icassp)}\ (\BPG~3145-3149).
\newblock
\begin{APACrefDOI} \doi{10.1109/ICASSP39728.2021.9414300} \end{APACrefDOI}
\PrintBackRefs{\CurrentBib}

\bibitem [\protect \citeauthoryear {%
Bouneffouf%
, Rish%
, Cecchi%
\BCBL {}\ \BBA {} Féraud%
}{%
Bouneffouf%
\ \protect \BOthers {.}}{%
{\protect \APACyear {2017}}%
}]{%
bouneffouf2017context}
\APACinsertmetastar {%
bouneffouf2017context}%
\begin{APACrefauthors}%
Bouneffouf, D.%
, Rish, I.%
, Cecchi, G.%
\BCBL {}\ \BBA {} Féraud, R.%
\end{APACrefauthors}%
\unskip\
\newblock
\APACrefYearMonthDay{2017}{}{}.
\newblock
{\BBOQ}\APACrefatitle {Context Attentive Bandits: Contextual Bandit with
  Restricted Context} {Context attentive bandits: Contextual bandit with
  restricted context}.{\BBCQ}
\newblock
\BIn{} \APACrefbtitle {International Joint Conference on Artificial
  Intelligence, {IJCAI-17}} {International joint conference on artificial
  intelligence, {IJCAI-17}}\ (\BPGS\ 1468--1475).
\newblock
\begin{APACrefURL} \url{https://doi.org/10.24963/ijcai.2017/203}
  \end{APACrefURL}
\newblock
\begin{APACrefDOI} \doi{10.24963/ijcai.2017/203} \end{APACrefDOI}
\PrintBackRefs{\CurrentBib}

\bibitem [\protect \citeauthoryear {%
Chapelle%
\ \BBA {} Li%
}{%
Chapelle%
\ \BBA {} Li%
}{%
{\protect \APACyear {2011}}%
}]{%
Chapelle2011}
\APACinsertmetastar {%
Chapelle2011}%
\begin{APACrefauthors}%
Chapelle, O.%
\BCBT {}\ \BBA {} Li, L.%
\end{APACrefauthors}%
\unskip\
\newblock
\APACrefYearMonthDay{2011}{}{}.
\newblock
{\BBOQ}\APACrefatitle {An {Empirical} {Evaluation} of {Thompson} {Sampling}}
  {An {Empirical} {Evaluation} of {Thompson} {Sampling}}.{\BBCQ}
\newblock
\BIn{} J.~Shawe-Taylor, R\BPBI S.~Zemel, P\BPBI L.~Bartlett, F.~Pereira\BCBL
  {}\ \BBA {} K\BPBI Q.~Weinberger\ (\BEDS), \APACrefbtitle {Neural Information
  Processing} {Neural information processing}\ (\BPGS\ 2249--2257).
\newblock
\APACaddressPublisher{}{Curran Associates, Inc.}
\newblock
\begin{APACrefURL}
  \url{http://papers.nips.cc/paper/4321-an-empirical-evaluation-of-thompson-sampling.pdf}
  \end{APACrefURL}
\PrintBackRefs{\CurrentBib}

\bibitem [\protect \citeauthoryear {%
Chu%
, Li%
, Reyzin%
\BCBL {}\ \BBA {} Schapire%
}{%
Chu%
\ \protect \BOthers {.}}{%
{\protect \APACyear {2011}}%
}]{%
Chu2011}
\APACinsertmetastar {%
Chu2011}%
\begin{APACrefauthors}%
Chu, W.%
, Li, L.%
, Reyzin, L.%
\BCBL {}\ \BBA {} Schapire, R.%
\end{APACrefauthors}%
\unskip\
\newblock
\APACrefYearMonthDay{2011}{11--13 Apr}{}.
\newblock
{\BBOQ}\APACrefatitle {Contextual Bandits with Linear Payoff Functions}
  {Contextual bandits with linear payoff functions}.{\BBCQ}
\newblock
\BIn{} G.~Gordon, D.~Dunson\BCBL {}\ \BBA {} M.~Dudík\ (\BEDS), (\BVOL~15,
  \BPGS\ 208--214).
\newblock
\APACaddressPublisher{Fort Lauderdale, FL, USA}{JMLR Workshop and Conference
  Proceedings}.
\newblock
\begin{APACrefURL} \url{http://proceedings.mlr.press/v15/chu11a.html}
  \end{APACrefURL}
\PrintBackRefs{\CurrentBib}

\bibitem [\protect \citeauthoryear {%
Garivier%
\ \BBA {} Moulines%
}{%
Garivier%
\ \BBA {} Moulines%
}{%
{\protect \APACyear {2011}}%
}]{%
Garivier2011}
\APACinsertmetastar {%
Garivier2011}%
\begin{APACrefauthors}%
Garivier, A.%
\BCBT {}\ \BBA {} Moulines, E.%
\end{APACrefauthors}%
\unskip\
\newblock
\APACrefYearMonthDay{2011}{}{}.
\newblock
{\BBOQ}\APACrefatitle {On Upper-Confidence Bound Policies for Switching Bandit
  Problems} {On upper-confidence bound policies for switching bandit
  problems}.{\BBCQ}
\newblock
\BIn{} J.~Kivinen, C.~Szepesv{\'a}ri, E.~Ukkonen\BCBL {}\ \BBA {} T.~Zeugmann\
  (\BEDS), \APACrefbtitle {Algorithmic Learning Theory} {Algorithmic learning
  theory}\ (\BPGS\ 174--188).
\newblock
\APACaddressPublisher{Berlin, Heidelberg}{Springer Berlin Heidelberg}.
\PrintBackRefs{\CurrentBib}

\bibitem [\protect \citeauthoryear {%
Gittins%
}{%
Gittins%
}{%
{\protect \APACyear {1979}}%
}]{%
Gittins1979}
\APACinsertmetastar {%
Gittins1979}%
\begin{APACrefauthors}%
Gittins, J.%
\end{APACrefauthors}%
\unskip\
\newblock
\APACrefYearMonthDay{1979}{}{}.
\newblock
{\BBOQ}\APACrefatitle {Bandit processes and dynamic allocation indices} {Bandit
  processes and dynamic allocation indices}.{\BBCQ}
\newblock
\APACjournalVolNumPages{Journal of the royal statistical society series
  b-methodological}{41}{}{148-164}.
\PrintBackRefs{\CurrentBib}

\bibitem [\protect \citeauthoryear {%
Huo%
\ \BBA {} Fu%
}{%
Huo%
\ \BBA {} Fu%
}{%
{\protect \APACyear {2017}}%
}]{%
Huo2017}
\APACinsertmetastar {%
Huo2017}%
\begin{APACrefauthors}%
Huo, X.%
\BCBT {}\ \BBA {} Fu, F.%
\end{APACrefauthors}%
\unskip\
\newblock
\APACrefYearMonthDay{2017}{}{}.
\newblock
{\BBOQ}\APACrefatitle {Risk-aware multi-armed bandit problem with application
  to portfolio selection} {Risk-aware multi-armed bandit problem with
  application to portfolio selection}.{\BBCQ}
\newblock
\APACjournalVolNumPages{Royal Society Open Science}{4}{11}{171377}.
\newblock
\begin{APACrefURL}
  \url{https://royalsocietypublishing.org/doi/abs/10.1098/rsos.171377}
  \end{APACrefURL}
\newblock
\begin{APACrefDOI} \doi{10.1098/rsos.171377} \end{APACrefDOI}
\PrintBackRefs{\CurrentBib}

\bibitem [\protect \citeauthoryear {%
Kaufmann%
, Korda%
\BCBL {}\ \BBA {} Munos%
}{%
Kaufmann%
\ \protect \BOthers {.}}{%
{\protect \APACyear {2012}}%
}]{%
Kaufmann2012}
\APACinsertmetastar {%
Kaufmann2012}%
\begin{APACrefauthors}%
Kaufmann, E.%
, Korda, N.%
\BCBL {}\ \BBA {} Munos, R.%
\end{APACrefauthors}%
\unskip\
\newblock
\APACrefYearMonthDay{2012}{}{}.
\newblock
{\BBOQ}\APACrefatitle {Thompson Sampling: An Asymptotically Optimal Finite-Time
  Analysis} {Thompson sampling: An asymptotically optimal finite-time
  analysis}.{\BBCQ}
\newblock
\BIn{} N\BPBI H.~Bshouty, G.~Stoltz, N.~Vayatis\BCBL {}\ \BBA {} T.~Zeugmann\
  (\BEDS), \APACrefbtitle {Algorithmic Learning Theory} {Algorithmic learning
  theory}\ (\BPGS\ 199--213).
\newblock
\APACaddressPublisher{Berlin, Heidelberg}{Springer Berlin Heidelberg}.
\PrintBackRefs{\CurrentBib}

\bibitem [\protect \citeauthoryear {%
{Kerkouche}%
, {Alami}%
, {Féraud}%
, {Varsier}%
\BCBL {}\ \BBA {} {Maillé}%
}{%
{Kerkouche}%
\ \protect \BOthers {.}}{%
{\protect \APACyear {2018}}%
}]{%
Kerkouche2018}
\APACinsertmetastar {%
Kerkouche2018}%
\begin{APACrefauthors}%
{Kerkouche}, R.%
, {Alami}, R.%
, {Féraud}, R.%
, {Varsier}, N.%
\BCBL {}\ \BBA {} {Maillé}, P.%
\end{APACrefauthors}%
\unskip\
\newblock
\APACrefYearMonthDay{2018}{}{}.
\newblock
{\BBOQ}\APACrefatitle {Node-based optimization of LoRa transmissions with
  Multi-Armed Bandit algorithms} {Node-based optimization of lora transmissions
  with multi-armed bandit algorithms}.{\BBCQ}
\newblock
\BIn{} \APACrefbtitle {2018 25th International Conference on Telecommunications
  (ICT)} {2018 25th international conference on telecommunications (ict)}\
  (\BPG~521-526).
\newblock
\begin{APACrefDOI} \doi{10.1109/ICT.2018.8464949} \end{APACrefDOI}
\PrintBackRefs{\CurrentBib}

\bibitem [\protect \citeauthoryear {%
Kocsis%
\ \BBA {} Szepesvári%
}{%
Kocsis%
\ \BBA {} Szepesvári%
}{%
{\protect \APACyear {2006}}%
}]{%
Kocis2006}
\APACinsertmetastar {%
Kocis2006}%
\begin{APACrefauthors}%
Kocsis, L.%
\BCBT {}\ \BBA {} Szepesvári, C.%
\end{APACrefauthors}%
\unskip\
\newblock
\APACrefYearMonthDay{2006}{}{}.
\newblock
{\BBOQ}\APACrefatitle {Discounted UCB} {Discounted ucb}.{\BBCQ}
\newblock
\APACjournalVolNumPages{2nd PASCAL Challenges Workshop}{}{}{}.
\newblock
\begin{APACrefURL} \url{https://www.lri.fr/~sebag/Slides/Venice/Kocsis.pdf}
  \end{APACrefURL}
\PrintBackRefs{\CurrentBib}

\bibitem [\protect \citeauthoryear {%
Lai%
\ \BBA {} Robbins%
}{%
Lai%
\ \BBA {} Robbins%
}{%
{\protect \APACyear {1985}}%
}]{%
Lai1985}
\APACinsertmetastar {%
Lai1985}%
\begin{APACrefauthors}%
Lai, T.%
\BCBT {}\ \BBA {} Robbins, H.%
\end{APACrefauthors}%
\unskip\
\newblock
\APACrefYearMonthDay{1985}{}{}.
\newblock
{\BBOQ}\APACrefatitle {Asymptotically efficient adaptive allocation rules}
  {Asymptotically efficient adaptive allocation rules}.{\BBCQ}
\newblock
\APACjournalVolNumPages{Advances in Applied Mathematics}{6}{1}{4 - 22}.
\newblock
\begin{APACrefURL}
  \url{http://www.sciencedirect.com/science/article/pii/0196885885900028}
  \end{APACrefURL}
\newblock
\begin{APACrefDOI} \doi{10.1016/0196-8858(85)90002-8} \end{APACrefDOI}
\PrintBackRefs{\CurrentBib}

\bibitem [\protect \citeauthoryear {%
Langford%
\ \BBA {} Zhang%
}{%
Langford%
\ \BBA {} Zhang%
}{%
{\protect \APACyear {2008}}%
}]{%
Langford2007}
\APACinsertmetastar {%
Langford2007}%
\begin{APACrefauthors}%
Langford, J.%
\BCBT {}\ \BBA {} Zhang, T.%
\end{APACrefauthors}%
\unskip\
\newblock
\APACrefYearMonthDay{2008}{}{}.
\newblock
{\BBOQ}\APACrefatitle {The Epoch-Greedy Algorithm for Multi-armed Bandits with
  Side Information} {The epoch-greedy algorithm for multi-armed bandits with
  side information}.{\BBCQ}
\newblock
\BIn{} J.~Platt, D.~Koller, Y.~Singer\BCBL {}\ \BBA {} S.~Roweis\ (\BEDS),
  \APACrefbtitle {Neural Information Processing} {Neural information
  processing}\ (\BVOL~20, \BPGS\ 817--824).
\newblock
\APACaddressPublisher{}{Curran Associates, Inc.}
\newblock
\begin{APACrefURL}
  \url{https://proceedings.neurips.cc/paper/2007/file/4b04a686b0ad13dce35fa99fa4161c65-Paper.pdf}
  \end{APACrefURL}
\PrintBackRefs{\CurrentBib}

\bibitem [\protect \citeauthoryear {%
Lei%
, Lu%
, Tewari%
\BCBL {}\ \BBA {} Murphy%
}{%
Lei%
\ \protect \BOthers {.}}{%
{\protect \APACyear {2017}}%
}]{%
Huitian2017}
\APACinsertmetastar {%
Huitian2017}%
\begin{APACrefauthors}%
Lei, H.%
, Lu, Y.%
, Tewari, A.%
\BCBL {}\ \BBA {} Murphy, S\BPBI A.%
\end{APACrefauthors}%
\unskip\
\newblock
\APACrefYearMonthDay{2017}{}{}.
\newblock
\APACrefbtitle {An Actor-Critic Contextual Bandit Algorithm for Personalized
  Mobile Health Interventions.} {An actor-critic contextual bandit algorithm
  for personalized mobile health interventions.}
\newblock
\APACaddressPublisher{}{arXiv}.
\newblock
\begin{APACrefURL} \url{https://arxiv.org/abs/1706.09090} \end{APACrefURL}
\newblock
\begin{APACrefDOI} \doi{10.48550/ARXIV.1706.09090} \end{APACrefDOI}
\PrintBackRefs{\CurrentBib}

\bibitem [\protect \citeauthoryear {%
Li%
, Chu%
, Langford%
\BCBL {}\ \BBA {} Schapire%
}{%
Li%
\ \protect \BOthers {.}}{%
{\protect \APACyear {2010}}%
}]{%
Li2010}
\APACinsertmetastar {%
Li2010}%
\begin{APACrefauthors}%
Li, L.%
, Chu, W.%
, Langford, J.%
\BCBL {}\ \BBA {} Schapire, R\BPBI E.%
\end{APACrefauthors}%
\unskip\
\newblock
\APACrefYearMonthDay{2010}{}{}.
\newblock
{\BBOQ}\APACrefatitle {A contextual-bandit approach to personalized news
  article recommendation} {A contextual-bandit approach to personalized news
  article recommendation}.{\BBCQ}
\newblock
\APACjournalVolNumPages{International conference on World wide web - WWW
  ’10}{}{}{}.
\newblock
\begin{APACrefURL} \url{http://dx.doi.org/10.1145/1772690.1772758}
  \end{APACrefURL}
\newblock
\begin{APACrefDOI} \doi{10.1145/1772690.1772758} \end{APACrefDOI}
\PrintBackRefs{\CurrentBib}

\bibitem [\protect \citeauthoryear {%
Mei%
, Xiao%
, Szepesvari%
\BCBL {}\ \BBA {} Schuurmans%
}{%
Mei%
\ \protect \BOthers {.}}{%
{\protect \APACyear {2020}}%
}]{%
Mei2020}
\APACinsertmetastar {%
Mei2020}%
\begin{APACrefauthors}%
Mei, J.%
, Xiao, C.%
, Szepesvari, C.%
\BCBL {}\ \BBA {} Schuurmans, D.%
\end{APACrefauthors}%
\unskip\
\newblock
\APACrefYearMonthDay{2020}{}{}.
\newblock
\APACrefbtitle {On the Global Convergence Rates of Softmax Policy Gradient
  Methods.} {On the global convergence rates of softmax policy gradient
  methods.}
\PrintBackRefs{\CurrentBib}

\bibitem [\protect \citeauthoryear {%
Ortner%
, Ryabko%
, Auer%
\BCBL {}\ \BBA {} Munos%
}{%
Ortner%
\ \protect \BOthers {.}}{%
{\protect \APACyear {2014}}%
}]{%
Ortner2014}
\APACinsertmetastar {%
Ortner2014}%
\begin{APACrefauthors}%
Ortner, R.%
, Ryabko, D.%
, Auer, P.%
\BCBL {}\ \BBA {} Munos, R.%
\end{APACrefauthors}%
\unskip\
\newblock
\APACrefYearMonthDay{2014}{}{}.
\newblock
{\BBOQ}\APACrefatitle {Regret bounds for restless Markov bandits} {Regret
  bounds for restless markov bandits}.{\BBCQ}
\newblock
\APACjournalVolNumPages{Theoretical Computer Science}{558}{}{62-76}.
\newblock
\begin{APACrefURL}
  \url{https://www.sciencedirect.com/science/article/pii/S030439751400704X}
  \end{APACrefURL}
\newblock
\APACrefnote{Algorithmic Learning Theory}
\newblock
\begin{APACrefDOI} \doi{10.1016/j.tcs.2014.09.026} \end{APACrefDOI}
\PrintBackRefs{\CurrentBib}

\bibitem [\protect \citeauthoryear {%
Prochaska%
, Redding%
, Evers%
\BCBL {}\ \protect \BOthers {.}}{%
Prochaska%
\ \protect \BOthers {.}}{%
{\protect \APACyear {2015}}%
}]{%
prochaska2015}
\APACinsertmetastar {%
prochaska2015}%
\begin{APACrefauthors}%
Prochaska, J\BPBI O.%
, Redding, C\BPBI A.%
, Evers, K\BPBI E.%
\BCBL {}\ \BOthersPeriod {.}\end{APACrefauthors}%
\unskip\
\newblock
\APACrefYearMonthDay{2015}{}{}.
\newblock
{\BBOQ}\APACrefatitle {The transtheoretical model and stages of change} {The
  transtheoretical model and stages of change}.{\BBCQ}
\newblock
\APACjournalVolNumPages{Health behavior: Theory, research, and
  practice}{97}{}{}.
\PrintBackRefs{\CurrentBib}

\bibitem [\protect \citeauthoryear {%
Schwartz%
, Bradlow%
\BCBL {}\ \BBA {} Fader%
}{%
Schwartz%
\ \protect \BOthers {.}}{%
{\protect \APACyear {2017}}%
}]{%
Schwartz2017}
\APACinsertmetastar {%
Schwartz2017}%
\begin{APACrefauthors}%
Schwartz, E.%
, Bradlow, E.%
\BCBL {}\ \BBA {} Fader, P.%
\end{APACrefauthors}%
\unskip\
\newblock
\APACrefYearMonthDay{2017}{04}{}.
\newblock
{\BBOQ}\APACrefatitle {Customer Acquisition via Display Advertising Using
  Multi-Armed Bandit Experiments} {Customer acquisition via display advertising
  using multi-armed bandit experiments}.{\BBCQ}
\newblock
\APACjournalVolNumPages{Marketing Science}{36}{}{}.
\newblock
\begin{APACrefDOI} \doi{10.1287/mksc.2016.1023} \end{APACrefDOI}
\PrintBackRefs{\CurrentBib}

\bibitem [\protect \citeauthoryear {%
Shen%
, Wang%
, Jiang%
\BCBL {}\ \BBA {} Zha%
}{%
Shen%
\ \protect \BOthers {.}}{%
{\protect \APACyear {2015}}%
}]{%
Shen2015}
\APACinsertmetastar {%
Shen2015}%
\begin{APACrefauthors}%
Shen, W.%
, Wang, J.%
, Jiang, Y\BHBI G.%
\BCBL {}\ \BBA {} Zha, H.%
\end{APACrefauthors}%
\unskip\
\newblock
\APACrefYearMonthDay{2015}{}{}.
\newblock
{\BBOQ}\APACrefatitle {Portfolio Choices with Orthogonal Bandit Learning}
  {Portfolio choices with orthogonal bandit learning}.{\BBCQ}
\newblock
\BIn{} \APACrefbtitle {International Conference on Artificial Intelligence}
  {International conference on artificial intelligence}\ (\BPG~974–980).
\newblock
\APACaddressPublisher{}{AAAI Press}.
\PrintBackRefs{\CurrentBib}

\bibitem [\protect \citeauthoryear {%
Sutton%
\ \BBA {} Barto%
}{%
Sutton%
\ \BBA {} Barto%
}{%
{\protect \APACyear {2018}}%
}]{%
Sutton2018}
\APACinsertmetastar {%
Sutton2018}%
\begin{APACrefauthors}%
Sutton, R\BPBI S.%
\BCBT {}\ \BBA {} Barto, A\BPBI G.%
\end{APACrefauthors}%
\unskip\
\newblock
\APACrefYear{2018}.
\newblock
\APACrefbtitle {Reinforcement Learning: An Introduction} {Reinforcement
  learning: An introduction}.
\newblock
\APACaddressPublisher{Cambridge, MA, USA}{A Bradford Book}.
\PrintBackRefs{\CurrentBib}

\bibitem [\protect \citeauthoryear {%
Thompson%
}{%
Thompson%
}{%
{\protect \APACyear {1933}}%
}]{%
Thompson1933}
\APACinsertmetastar {%
Thompson1933}%
\begin{APACrefauthors}%
Thompson, W\BPBI R.%
\end{APACrefauthors}%
\unskip\
\newblock
\APACrefYearMonthDay{1933}{}{}.
\newblock
{\BBOQ}\APACrefatitle {On the Likelihood that One Unknown Probability Exceeds
  Another in View of the Evidence of Two Samples} {On the likelihood that one
  unknown probability exceeds another in view of the evidence of two
  samples}.{\BBCQ}
\newblock
\APACjournalVolNumPages{Biometrika}{25}{3/4}{285--294}.
\newblock
\begin{APACrefURL} \url{http://www.jstor.org/stable/2332286} \end{APACrefURL}
\PrintBackRefs{\CurrentBib}

\bibitem [\protect \citeauthoryear {%
Villar%
, Bowden%
\BCBL {}\ \BBA {} Wason%
}{%
Villar%
\ \protect \BOthers {.}}{%
{\protect \APACyear {2015}}%
}]{%
Villar2015}
\APACinsertmetastar {%
Villar2015}%
\begin{APACrefauthors}%
Villar, S\BPBI S.%
, Bowden, J.%
\BCBL {}\ \BBA {} Wason, J.%
\end{APACrefauthors}%
\unskip\
\newblock
\APACrefYearMonthDay{2015}{}{}.
\newblock
{\BBOQ}\APACrefatitle {{{M}ulti-armed {B}andit {M}odels for the {O}ptimal
  {D}esign of {C}linical {T}rials: {B}enefits and {C}hallenges}}
  {{{M}ulti-armed {B}andit {M}odels for the {O}ptimal {D}esign of {C}linical
  {T}rials: {B}enefits and {C}hallenges}}.{\BBCQ}
\newblock
\APACjournalVolNumPages{Stat Sci}{30}{2}{199--215}.
\newblock
\begin{APACrefURL} \url{https://doi.org/10.1214/14-STS504} \end{APACrefURL}
\PrintBackRefs{\CurrentBib}

\bibitem [\protect \citeauthoryear {%
{Wang}%
\ \protect \BOthers {.}}{%
{Wang}%
\ \protect \BOthers {.}}{%
{\protect \APACyear {2019}}%
}]{%
Wang2019}
\APACinsertmetastar {%
Wang2019}%
\begin{APACrefauthors}%
{Wang}, Q.%
, {Zeng}, C.%
, {Zhou}, W.%
, {Li}, T.%
, {Iyengar}, S\BPBI S.%
, {Shwartz}, L.%
\BCBL {}\ \BBA {} {Grabarnik}, G\BPBI Y.%
\end{APACrefauthors}%
\unskip\
\newblock
\APACrefYearMonthDay{2019}{}{}.
\newblock
{\BBOQ}\APACrefatitle {Online Interactive Collaborative Filtering Using
  Multi-Armed Bandit with Dependent Arms} {Online interactive collaborative
  filtering using multi-armed bandit with dependent arms}.{\BBCQ}
\newblock
\APACjournalVolNumPages{IEEE Transactions on Knowledge and Data
  Engineering}{31}{8}{1569-1580}.
\newblock
\begin{APACrefDOI} \doi{10.1109/TKDE.2018.2866041} \end{APACrefDOI}
\PrintBackRefs{\CurrentBib}

\bibitem [\protect \citeauthoryear {%
Wen%
, Kveton%
, Valko%
\BCBL {}\ \BBA {} Vaswani%
}{%
Wen%
\ \protect \BOthers {.}}{%
{\protect \APACyear {2017}}%
}]{%
Wen2017}
\APACinsertmetastar {%
Wen2017}%
\begin{APACrefauthors}%
Wen, Z.%
, Kveton, B.%
, Valko, M.%
\BCBL {}\ \BBA {} Vaswani, S.%
\end{APACrefauthors}%
\unskip\
\newblock
\APACrefYearMonthDay{2017}{}{}.
\newblock
{\BBOQ}\APACrefatitle {Online Influence Maximization under Independent Cascade
  Model with Semi-Bandit Feedback} {Online influence maximization under
  independent cascade model with semi-bandit feedback}.{\BBCQ}
\newblock
\BIn{} I.~Guyon\ \BOthers {.}\ (\BEDS), \APACrefbtitle {Advances in Neural
  Information Processing Systems} {Advances in neural information processing
  systems}\ (\BVOL~30, \BPGS\ 3022--3032).
\newblock
\APACaddressPublisher{}{Curran Associates, Inc.}
\newblock
\begin{APACrefURL}
  \url{https://proceedings.neurips.cc/paper/2017/file/7137debd45ae4d0ab9aa953017286b20-Paper.pdf}
  \end{APACrefURL}
\PrintBackRefs{\CurrentBib}

\bibitem [\protect \citeauthoryear {%
Whittle%
}{%
Whittle%
}{%
{\protect \APACyear {1988}}%
}]{%
Whittle88}
\APACinsertmetastar {%
Whittle88}%
\begin{APACrefauthors}%
Whittle, P.%
\end{APACrefauthors}%
\unskip\
\newblock
\APACrefYearMonthDay{1988}{}{}.
\newblock
{\BBOQ}\APACrefatitle {Restless Bandits: Activity Allocation in a Changing
  World} {Restless bandits: Activity allocation in a changing world}.{\BBCQ}
\newblock
\APACjournalVolNumPages{Journal of Applied Probability}{25}{}{287--298}.
\newblock
\begin{APACrefURL} \url{http://www.jstor.org/stable/3214163} \end{APACrefURL}
\newblock
\begin{APACrefDOI} \doi{10.2307/3214163} \end{APACrefDOI}
\PrintBackRefs{\CurrentBib}

\bibitem [\protect \citeauthoryear {%
Yu%
\ \BBA {} Mannor%
}{%
Yu%
\ \BBA {} Mannor%
}{%
{\protect \APACyear {2009}}%
}]{%
Yu2009}
\APACinsertmetastar {%
Yu2009}%
\begin{APACrefauthors}%
Yu, J\BPBI Y.%
\BCBT {}\ \BBA {} Mannor, S.%
\end{APACrefauthors}%
\unskip\
\newblock
\APACrefYearMonthDay{2009}{}{}.
\newblock
{\BBOQ}\APACrefatitle {Piecewise-Stationary Bandit Problems with Side
  Observations} {Piecewise-stationary bandit problems with side
  observations}.{\BBCQ}
\newblock
\BIn{} \APACrefbtitle {Proceedings of the 26th Annual International Conference
  on Machine Learning} {Proceedings of the 26th annual international conference
  on machine learning}\ (\BPG~1177–1184).
\newblock
\APACaddressPublisher{New York, NY, USA}{Association for Computing Machinery}.
\newblock
\begin{APACrefURL} \url{https://doi.org/10.1145/1553374.1553524}
  \end{APACrefURL}
\newblock
\begin{APACrefDOI} \doi{10.1145/1553374.1553524} \end{APACrefDOI}
\PrintBackRefs{\CurrentBib}

\bibitem [\protect \citeauthoryear {%
Yun%
, Nam%
, Mo%
\BCBL {}\ \BBA {} Shin%
}{%
Yun%
\ \protect \BOthers {.}}{%
{\protect \APACyear {2017}}%
}]{%
Yun2017}
\APACinsertmetastar {%
Yun2017}%
\begin{APACrefauthors}%
Yun, S\BHBI Y.%
, Nam, J\BPBI H.%
, Mo, S.%
\BCBL {}\ \BBA {} Shin, J.%
\end{APACrefauthors}%
\unskip\
\newblock
\APACrefYearMonthDay{2017}{}{}.
\newblock
{\BBOQ}\APACrefatitle {{Contextual Multi-armed Bandits under Feature
  Uncertainty}} {{Contextual Multi-armed Bandits under Feature
  Uncertainty}}.{\BBCQ}
\newblock
\APACjournalVolNumPages{}{}{}{1--25}.
\newblock
\begin{APACrefURL} \url{http://arxiv.org/abs/1703.01347} \end{APACrefURL}
\newblock
\begin{APACrefDOI} \doi{10.2172/1345927} \end{APACrefDOI}
\PrintBackRefs{\CurrentBib}

\bibitem [\protect \citeauthoryear {%
Zhou%
, Xiong%
, Chen%
\BCBL {}\ \BBA {} Gao%
}{%
Zhou%
\ \protect \BOthers {.}}{%
{\protect \APACyear {2021}}%
}]{%
Zhou2021}
\APACinsertmetastar {%
Zhou2021}%
\begin{APACrefauthors}%
Zhou, X.%
, Xiong, Y.%
, Chen, N.%
\BCBL {}\ \BBA {} Gao, X.%
\end{APACrefauthors}%
\unskip\
\newblock
\APACrefYearMonthDay{2021}{}{}.
\newblock
\APACrefbtitle {Regime Switching Bandits.} {Regime switching bandits.}
\newblock
\begin{APACrefURL} \url{https://arxiv.org/abs/2001.09390} \end{APACrefURL}
\PrintBackRefs{\CurrentBib}

\end{thebibliography}

\renewcommand{\thetable}{A.\arabic{table}}

\newpage
\appendix
\section{Derivation of the Gradient Bandit Learning Dynamics}\label{sec:ap}

We iterate the update equations for our gradient bandit formulation:

\begin{align}
\label{eq:gb1_a} &H_{i,\pi_{i}} \gets H_{i,\pi_{i}} + \delta_R\big(r_{i,a_{i}}-pb_{i, \pi_{i}}\big),\\
\label{eq:gb2_a} &H_{i,\neg \pi_{i}} \gets H_{i, \neg \pi_{i}} + \delta_R \big(1 - 2r_{i,a_{i}}\big) \big(1 - pb_{i,\pi_{i}}\big).
\end{align}

First, we are looking for the change in probability $pb$ of choosing a policy depending on the change in weights of the softmax policy and the change of the weights $H_\pi$ over time, that is:

\begin{equation}\label{eq:ptotdiff}
    \frac{d pb_{i,\pi}}{dt} = \frac{d pb_{i,\pi}}{dH_{i,\pi}} \times \frac{d H_{i,\pi}}{dt}.
\end{equation}

The derivative of the softmax policy with respect to weights $H_{i,\pi}$ is simply~\cite{Sutton2018}:

\begin{equation}\label{eq:diffp}
    \frac{d pb_{i,k}}{dH_{i,j}} = pb_i(I_{k,j} - pb_j),
\end{equation}
where $I$ is the indicator function being $1$ if $k=j$ and $0$ otherwise. We now derive the expression $\frac{d H_{i,\pi}}{dt}$. For the derivation, we assume that the gradient bandit needs to decide between two policies and assume Bernoulli distributed rewards. Furthermore, we omit the subscript $i$ for ease of notation. Thus, given the update equations above, we have to consider the following four cases:

\begin{flalign*}
Case\;1: H_{\pi_{t}} = \pi_0 \land r_{t,a_{t}} = 1: \\
H_{\pi_0} &\gets H_{\pi_0} + \delta_R\big(1-pb_{\pi_0}\big)\\
H_{\pi_1} &\gets H_{\pi_1} - \delta_R\big(1-pb_{\pi_0}\big)&&
\end{flalign*}
\begin{flalign*}
Case\;2: H_{\pi_{t}} = \pi_0 \land r_{t,a_{t}} = 0: \\
H_{\pi_0} &\gets H_{\pi_0} - \delta_Rpb_{\pi_0}\\
H_{\pi_1} &\gets H_{\pi_1} + \delta_R\big(1-pb_{\pi_0}\big)&&
\end{flalign*}

\begin{flalign*}
Case\;3: H_{\pi_{t}} = \pi_1 \land r_{t,a_{t}} = 1: \\
H_{\pi_0} &\gets H_{\pi_0} - \delta_Rpb_{\pi_0}\\
H_{\pi_1} &\gets H_{\pi_1} + \delta_Rpb_{\pi_0}&&
\end{flalign*}

\begin{flalign*}
Case\;4: H_{\pi_{t}} = \pi_1 \land r_{t,a_{t}} = 0: \\
H_{\pi_0} &\gets H_{\pi_0} + \delta_Rpb_{\pi_0}\\
H_{\pi_1} &\gets H_{\pi_1} - \delta_R\big(1-pb_{\pi_0}\big)&&
\end{flalign*}

Note the second term on the RHS of each of these update equations can be interpreted as a case-dependent derivative of the respective weights. Note that we have expressed all probabilities in terms of one policy $\pi_0$, which we can do since $pb_1:= 1-pb_0$.

First note that for each case, given average reward $r_{\pi_0}$ for policy $\pi_0$ and average reward $r_{\pi_1}$ for policy $\pi_1$, we construct an expectation over derivatives by a weighted average for each of the cases above. We start with derivative $\frac{d H_{\pi_0}}{d t}$. If $\pi_t = \pi_0$, as a first step, consider the weighted sum of all the update equations for $H_{\pi_0}$ where $r_{t,a_t} = 1 $ or  $r_{t,a_t} = 0$. The weights in the sum are determined by the average reward $r_{\pi_0}$. For $r_{t,a_t} = 1$ we weight the update with $r_{\pi_0}$ and for $r_{t,a_t} = 0$ with $(1-r_{\pi_0})$ leading to the partial term:

\begin{equation}\label{eq:p1div}
r_{\pi_0} \delta_R\big(1-pb_{\pi_0}\big)  -  (1-r_{\pi_0}) \delta_Rpb_{\pi_0}.
\end{equation}

Now $H_{\pi_0}$ is also updated when $\pi_t = \pi_1$. The weights are now determined by the average reward $r_{\pi_1}$, thus we get for the second partial term:

\begin{equation}\label{eq:p2div}
-r_{\pi_1}\delta_R pb_{\pi_0} + (1-r_{\pi_1})\delta_R pb_{\pi_0}.
\end{equation}

For the final result, we need to consider the probability of these two events happening. This is determined by probabilities $pb_{\pi_0}$ and $pb_{\pi_1}$. We multiply equations \ref{eq:p1div} and \ref{eq:p2div} with the respective probabilities getting:

\begin{equation}\label{eq:p0divres}
 pb_{\pi_0} \bigg(r_{\pi_0} \delta_R\big(1-pb_{\pi_0}\big)  -  (1-r_{\pi_0}) \delta_Rpb_{\pi_0}\bigg),\\
 (1- pb_{\pi_0})\bigg(-r_{\pi_1}\delta_Rpb_{\pi_0} + (1-r_{\pi_1})\delta_Rpb_{\pi_0}\bigg),
\end{equation}

where we substituted $pb_1 = (1- pb_0)$. $\frac{d H_{\pi_1}}{d t}$ can be derived in a similar fashion. We then have for the derivatives of the weights $H_\pi$:

\begin{multline}\label{eq:diffh0}
\frac{dH_{\pi_0}}{dt} = pb_{\pi_0} \big(r_{\pi_0} \delta_R\big(1-pb_{\pi_0}\big)  -  (1-r_{\pi_0}) \delta_Rpb_{\pi_0}\big) \\
+ (1- pb_{\pi_0})\big(-r_{\pi_1}\delta_Rpb_{\pi_0} + (1-r_{\pi_1})\delta_Rpb_{\pi_0}\big),
\end{multline}
\begin{multline}\label{eq:diffh1}
\frac{dH_{\pi_1}}{dt} = pb_{\pi_0}\big(- r_{\pi_0}\delta_R\big(1-pb_{\pi_0}\big) +  (1-r_{\pi_0})\delta_R\big(1-pb_{\pi_0}\big)\big) \\
+ (1- pb_{\pi_0})\big( r_{\pi_1}\delta_Rpb_{\pi_0} - (1-r_{\pi_1})\delta_R\big(1-pb_{\pi_0}\big)\big).
\end{multline}

We now derive the final expression in equation \ref{eq:ptotdiff}, we focus on $\frac{dpb_{\pi_0}}{dt}$, since the change in probability of $\pi_1$ follows directly by the fact $\frac{dpb_{\pi_1}}{dt} = -\frac{dpb_{\pi_0}}{dt}$. In order to describe $\frac{d pb_{\pi_0}}{dt}$ we need to consider the total differential:

\begin{equation}\label{eq:difftotal}
    \frac{d pb_{\pi_0}}{dt} = \frac{d pb_{\pi_0}}{dH_{\pi_0}} \times \frac{d H_{\pi_0}}{dt} + \frac{d pb_{\pi_0}}{dH_{\pi_1}} \times \frac{d H_{\pi_1}}{dt}.
\end{equation}

Putting equations \ref{eq:diffp}, \ref{eq:diffh0} and \ref{eq:diffh1} in \ref{eq:difftotal}, we get after some simplification:

\begin{equation}
\frac{d p^*(t)}{d t} = \delta_R
p^*(t)(\Delta_R - r^* + p^*(t)(2 p^*(t) - 3)(\Delta_R p^*(t) - r^* + 1) + 1),
\end{equation}

where $\Delta_R = r_{\pi_0} - r_{\pi_1} > 0$, $r^* = r_{\pi_0}$ and $p^*(t) = pb_{\pi_0}(t)$.

\end{document}